\documentclass{article}

    \PassOptionsToPackage{numbers, square, sort}{natbib}

\usepackage[final]{neurips_2021}

\usepackage[utf8]{inputenc} 
\usepackage[T1]{fontenc}    
\usepackage{hyperref}       
\usepackage{url}            
\usepackage{booktabs}       
\usepackage{amsfonts}       
\usepackage{nicefrac}       
\usepackage{microtype}      
\usepackage{xcolor}         

\usepackage{graphicx}
\usepackage{amssymb}
\usepackage{amsthm}
\usepackage{amsmath}
\usepackage{soul}
\usepackage{booktabs}
\usepackage{bbm}
\usepackage{listings}
\usepackage{tabularx}
\usepackage{enumitem}

\usepackage[T1]{fontenc}
\usepackage[utf8]{inputenc}
\usepackage{authblk}

\usepackage[caption=false]{subfig}
\graphicspath{{figures/}}

\definecolor{mygreen}{rgb}{0,0.6,0}
\definecolor{mygray}{rgb}{0.5,0.5,0.5}
\definecolor{mymauve}{rgb}{0.58,0,0.82}

\hypersetup{
    colorlinks,
    linkcolor={red!50!black},
    citecolor={blue!50!black},
    urlcolor={blue!80!black}
}

\title{\papertitle}

\author[1,3]{Victor Zhong\thanks{Corresponding author Victor Zhong~\url{vzhong@cs.washington.edu}}}
\author[2]{Austin W. Hanjie}
\author[3]{Sida I. Wang}
\author[2]{Karthik Narasimhan}
\author[1,3]{Luke Zettlemoyer}
\affil[1]{Department of Computer Science, University of Washington}
\affil[2]{Department of Computer Science, Princeton University}
\affil[3]{Facebook AI Research}

\renewcommand\footnotemark{}

\begin{document}
\newcommand{\tocite}[1]{{[\hl{CITE: #1]}}}
\newcommand{\todo}[1]{{[\hl{TODO: #1}]}}
\newcommand{\victor}[1]{{[\hl{VICTOR: #1}]}}
\newcommand{\luke}[1]{{[\hl{LUKE: #1}]}}
\newcommand{\karthik}[1]{{[\hl{KARTHIK: #1}]}}
\newcommand{\austin}[1]{{[\hl{AUSTIN: #1}]}}
\newcommand{\sida}[1]{{[\hl{SIDA: #1}]}}
\newcommand{\camrdy}[1]{{#1}}

\newcommand{\benchmarkname}{{SILG}}
\newcommand{\papertitle}{{\benchmarkname: The Multi-environment Symbolic Interactive Language Grounding Benchmark}}

\newcommand{\modelname}{{Symbolic Interactive Reader}}
\newcommand{\modelnameshort}{{SIR}}

\newcommand{\caveat}{{$^*$}}
\newcommand{\sfunction}[1]{\textsf{\textsc{#1}}}

\newenvironment{minipeqn}[1][]{\begin{minipage}[#1]{.45\textwidth}\begin{equation}}{\end{equation}\end{minipage}}

\newcommand{\real}[1]{{\mathbb{R}^{{#1}}}}
\newcommand{\embed}[1]{{\rm emb} \left( {{#1}} \right)}
\newcommand{\BiLSTM}[2]{{\rm BiLSTM}_{#1} \left( {{#2}} \right)}
\newcommand{\LSTM}[1]{{\rm LSTM} \left( {{#1}} \right)}
\newcommand{\softmax}[1]{{\rm softmax} \left( {{#1}} \right)}
\newcommand{\linear}[2]{{\rm linear}_{#1} \left( {{#2}} \right)}
\newcommand{\concat}[1]{{\left[ #1 \right]}}
\newcommand{\flatten}[1]{{\rm flatten} {#1}}
\newcommand{\attend}[1]{{\rm attend} \left( {{#1}} \right)}
\newcommand{\selfattn}[2]{{\rm weightave}_{#1} \left( {{#2}} \right)}
\newcommand{\film}{{${\rm FiLM}^2$}}
\newcommand{\fsum}[1]{{\rm sum} \left( {{#1}} \right)}

\newcommand{\baseline}{B}
\newcommand{\policy}{{Y}}
\newcommand{\mlp}[2]{{\rm MLP}_{#1} \left( {{#2}} \right)}

\newcommand{\lstmstate}{{S}}

\newcommand{\repfinal}{{H}}
\newcommand{\repcmd}{{G}}
\newcommand{\repdirection}{{I}}
\newcommand{\repworld}{{U}}
\newcommand{\repdynamics}{{D}}
\newcommand{\repnondynamics}{{N}}

\newcommand{\cmd}{{Q}}
\newcommand{\direction}{{E}}

\newcommand{\world}{{X}}
\newcommand{\worldheight}{{h}}
\newcommand{\worldwidth}{{w}}
\newcommand{\dencode}{{q}}

\newcommand{\dynamics}{{T}}
\newcommand{\ldynamics}{{l}}
\newcommand{\nondynamics}{{R}}
\newcommand{\lnondynamics}{{m}}
\newcommand{\nnondynamics}{{n}}

\newcommand{\demb}{{d}}
\newcommand{\drnn}{{r}}
\newcommand{\lent}{{k}}

\newcommand{\relpos}{{Z}}

\newcommand{\sanondynamics}{{C}}
\newcommand{\attnnondynamics}{{A}}

\newcommand{\weight}{W}
\newcommand{\vectorweight}{w}
\newcommand{\bias}{b}
\newcommand{\scalarbias}{b}
\newcommand{\vectorgamma}{\gamma}
\newcommand{\vectorbeta}{\beta}
\newcommand{\matrixgamma}{\Gamma}
\newcommand{\matrixbeta}{B}
\newcommand{\len}{l}
\newcommand{\filmfeat}{x}
\newcommand{\filmmatrixfeat}{X}
\newcommand{\filmout}{V}
\newcommand{\filmsumm}{s}
\newcommand{\rnnhid}{H}
\newcommand{\vectorrnnhid}{h}
\newcommand{\attnscore}{a^\prime}
\newcommand{\scalarattnscore}{a^\prime}
\newcommand{\normattnscore}{a}
\newcommand{\scalarnormattnscore}{a}
\newcommand{\attncontext}{{c}}
\newcommand{\outrep}{\filmsumm^{(\mathrm{last})}}
\newcommand{\outtransrep}{o}

\newcommand{\conv}{{\rm Conv}}
\newcommand{\relu}{{\rm ReLU}}
\newcommand{\maxpool}{{\rm MaxPool}}
%
%

\newcommand{\vtext}{{\rm text}}
\newcommand{\vwiki}{{\rm doc}}
\newcommand{\vvis}{{\rm vis}}
\newcommand{\vobs}{{\rm obs}}
\newcommand{\vtask}{{\rm goal}}
\newcommand{\vinv}{{\rm inv}}
\newcommand{\vemb}{{\rm emb}}
\newcommand{\vhid}{{\rm hid}}
\newcommand{\vpos}{{\rm pos}}
\newcommand{\vini}{{\rm ini}}

\maketitle

\begin{abstract}
Existing work in language grounding typically study single environments.
How do we build unified models that apply across multiple environments?
We propose the multi-environment Symbolic Interactive Language Grounding benchmark (\benchmarkname), which unifies a collection of diverse grounded language learning environments under a common interface.
\benchmarkname~consists of grid-world environments that require generalization to new dynamics, entities, and partially observed worlds (RTFM, Messenger, NetHack), as well as symbolic counterparts of visual worlds that require interpreting rich natural language with respect to complex scenes (ALFWorld, Touchdown).
Together, these environments provide diverse grounding challenges in richness of observation space, action space, language specification, and plan complexity.
In addition, we propose the first shared model architecture for RL on these environments, and evaluate recent advances such as egocentric local convolution, recurrent state-tracking, entity-centric attention, and pretrained LM using~\benchmarkname.
Our shared architecture achieves comparable performance to environment-specific architectures.
Moreover, we find that many recent modelling advances do not result in significant gains on environments other than the one they were designed for.
This highlights the need for a multi-environment benchmark.
Finally, the best models significantly underperform humans on~\benchmarkname, which suggests ample room for future work.
We hope~\benchmarkname~enables the community to quickly identify new methodologies for language grounding that generalize to a diverse set of environments and their associated challenges.
\end{abstract}


\section{Introduction}

An ideal language-conditioned agent should interpret language in diverse environments with varying observation space, action space, language, and plan complexity.
However, existing language-grounding literature typically focuses on single environments, and proposes methodological contributions specific to those environments~\citep{luketina2019survey,tellex2020robots}.
In order to determine which contributions are environment-specific and which apply across multiple environments, it is critical to develop universal models that can be easily evaluated in many different settings.

To facilitate this research, we present the multi-environment Symbolic Interactive Language Grounding Benchmark (\benchmarkname).
We focus on symbolic environments with semantic symbols instead of raw visual observations for efficiency, interpretability, and emphasis on abstractions over perception.
\benchmarkname~consists of diverse environments including grid-worlds RTFM~\citep{zhong2020rtfm}, Messenger~\citep{hanjie21grounding}, and NetHack~\citep{kuttler2020nethack}, which require generalization to new dynamics (i.e.~how entities behave), entity references, and partially observed worlds. 
\benchmarkname~also contains symbolic counterparts of visual grounding environments ALFRED~\citep{shridhar2020alfred} and Touchdown~\citep{chen2018touchdown}
, which require interpreting rich natural language in complex scenes.
For the former, we use its textual variant ALFWorld~\citep{shridhar2021alfworld}.
For the latter, we create SymTD by applying object segmentation to Touchdown panoramas.
Despite significant implementation differences, we unify these environments under a common interface in~\benchmarkname, so that one can easily develop and evaluate language grounded RL methods across all of these challenges.

\benchmarkname~environments present a variety of unique grounding challenges in the richness of the observation space, action space, language specification, and plan complexity.
We quantify these challenges and additionally analyze the success rate and lengths of expert playthroughs.
For visual grounding environments, we show symbolic variants (ALFWorld and SymTD) facilitate faster learning and result in policies that transfer to their visual counterparts. \camrdy{While a unified model may not outperform specialized models engineered for specific environments, it can be helpful to understand whether particular modelling innovations are environment specific or more general techniques. Furthermore, while the challenges in each environment are very different, we want to encourage the development of unified architectures and approaches that can scale across many language grounding tasks.}

In addition to~\benchmarkname, we propose the~\modelname~(\modelnameshort), the first shared model architecture for these environments.
We combine~\modelnameshort~with several recent advances in language-conditioned RL, including~\film~\citep{zhong2020rtfm}, egocentric local convolution~\citep{hill2019environmental}, recurrent state-tracking~\citep{kuttler2020nethack}, entity-centric attention~\citep{hanjie21grounding}, and large pretrained LMs~\citep{hill2020human}.
On most environments, ~\modelnameshort~achieves comparable performance to methods designed specifically for single environments.
In addition, we find that many recent  advances do not result in significant gains on environments other than the one they were designed for.
This highlights the need for a multi-environment benchmark.
Finally, the best models significantly underperform humans on~\benchmarkname~(10-85\% depending on environment), which suggests ample room for modelling improvements that generalize across environments.

In summary, we (1) combine five language-grounding environments under the same interface to evaluate language grounded RL methods across diverse grounding challenges, (2) present the first shared model architecture for these environments, and (3) analyze recent modelling contributions across these environments.
We hope~\benchmarkname~enables the community to quickly identify new models and learning algorithms that generalize to a diverse set of environments and their associated challenges.
The code for~\benchmarkname~is available at~\url{https://github.com/vzhong/silg}.

\begin{figure}
    \centering
    \includegraphics[width=0.90\linewidth]{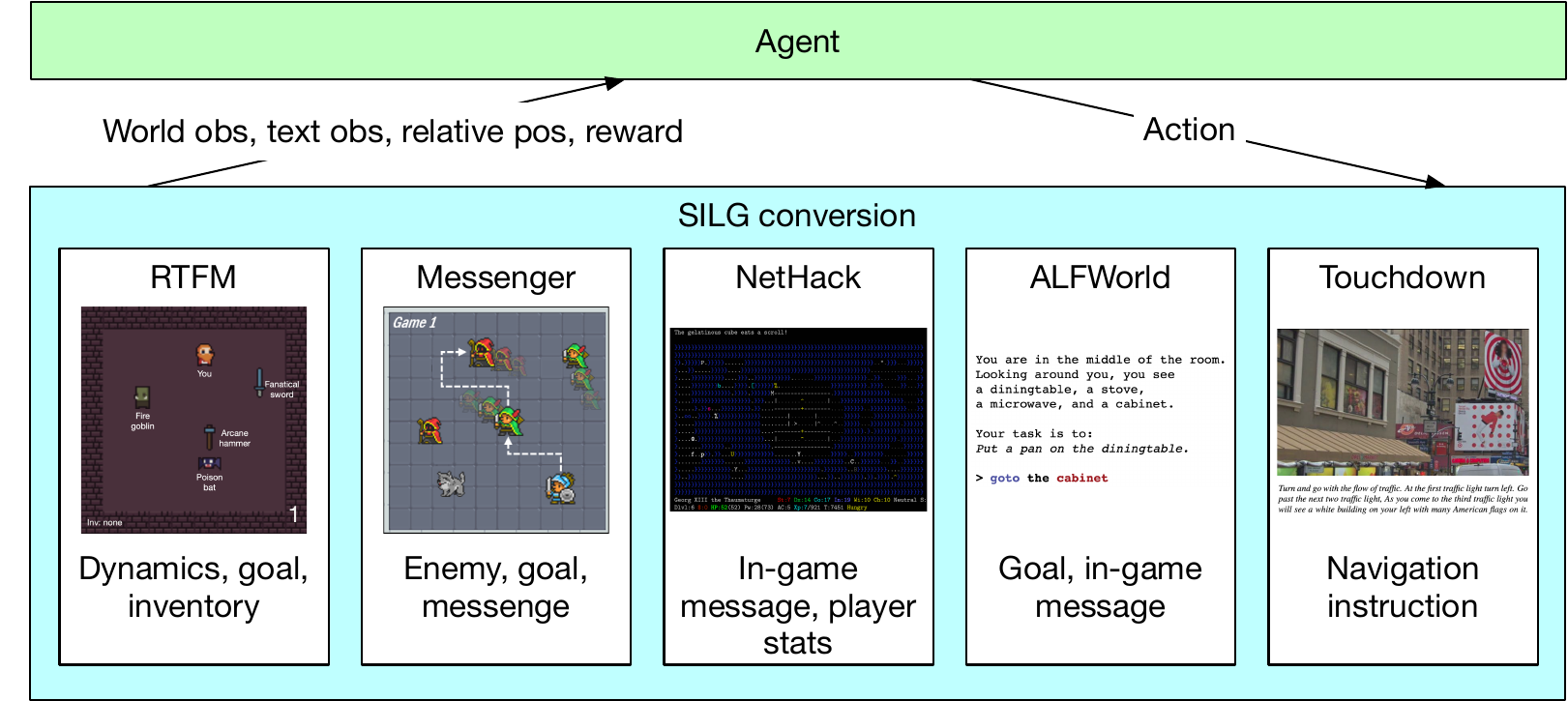}
    \vspace{-0.1in}
    \caption{Environments included in SILG. The world observations and text fields are shown for each environment. Detailed examples are in Appendix~\ref{app:experts}.}
    \label{fig:environments}
    \vspace{-0.25in}
\end{figure}

\begin{table}[t]
\footnotesize
\centering
\caption{\benchmarkname~statistics.
``dynamics'' are high level rules dictating behaviour of entities.
``Ref hops'' are number of intra-text references the agent must resolve to determine correct course of action.
Messenger and SymTD text are human-written instead of procedurally generated. Distinctive properties are~\textbf{bold}.
}
\label{tab:silg_stats}
\begin{tabularx}{\linewidth}{@{}XXXXXX@{}}
\toprule
 & RTFM & Messenger & SILGNetHack & ALFWorld & SymTD \\ \midrule
\textbf{Action space} & 5 fixed & 5 fixed & 23 fixed & \textbf{50+ choices} & 1-5 choices \\
\midrule
\textbf{State space} & \begin{tabular}[c]{@{}l@{}}$6\times6$ grid\\ 5 entities\end{tabular} & \begin{tabular}[c]{@{}l@{}}$10\times10$ grid\\ 14 entities\end{tabular} & \begin{tabular}[c]{@{}l@{}}$21\times79$\\ \textbf{partial obs}\end{tabular} & \begin{tabular}[c]{@{}l@{}}102 nodes\\ 191 entities\end{tabular}  & \begin{tabular}[c]{@{}l@{}}\textbf{29.6k complex}\\\textbf{panoramas}\end{tabular} \\
\midrule
\textbf{Mean text len} & 31 words & 30 words & 9 words & \textbf{100 words} & \textbf{90 words} \\ \midrule
\textbf{Vocab size} & 262 words & 595 words & $\sim$100 words & 1237 words & \textbf{4999 words} \\
\midrule
\textbf{Generalization} & \textbf{new dynamics} & \textbf{new dynamics} & new layouts & \begin{tabular}[c]{@{}l@{}}new instr\\ new layouts\end{tabular} & new instr \\
\midrule
\textbf{Ref hops} & \textbf{6 hops} & 3 hops & 1 hop & $\sim$4 hops & $\sim$\textbf{7 hops} \\
\midrule
\textbf{Human win} \% & 100\% & 100\% & 78.1\% & \begin{tabular}[c]{@{}l@{}}100\% new instr\\ 100\% +layouts\end{tabular} & 61.5\% \\
\midrule
\textbf{Human \# steps} & 6.0 steps & 2.2 steps & 34.4 steps & \begin{tabular}[c]{@{}l@{}}7.8 steps new instr\\ 9.6 steps +layouts\end{tabular} & 33.6 steps\\
\midrule
\camrdy{\textbf{Env FPS}} & 240 & 1627 & 439 & 7 & 779\\
\midrule
\textbf{Key challenge} &
\begin{tabular}[c]{@{}l@{}}multi-step\\ reasoning\end{tabular} &
\begin{tabular}[c]{@{}l@{}}adversarial\\ generalization\end{tabular} &
partial obs &
\begin{tabular}[c]{@{}l@{}}large\\action space\end{tabular} &
\begin{tabular}[c]{@{}l@{}}complex\\ language\end{tabular}\\
\bottomrule
\end{tabularx}
\vspace{-0.2in}
\end{table}

\section{\benchmarkname~Environments}
\label{sec:benchmark}

\benchmarkname~contains five language-grounding environments including both grid-worlds (RTFM, Messenger, SILGNetHack) and symbolic counterparts of 3D-visual worlds (ALFWorld, SymTD).
While all involve agents situated in interactive worlds, each presents unique challenges in richness of observation space, action space, language specification, and plan complexity. 
Table~\ref{tab:silg_stats} quantifies their theoretical complexity along these dimensions as well as empirical complexity using expert playthroughs.\footnote{For each environment, an expert plays as many episodes as necessary to learn about the game. We then record the playthroughs to compute the empirical win rate and trajectory length. More details in Appendix~\ref{app:experts}.}

\begin{figure}
    \centering
    \includegraphics[width=0.8\linewidth]{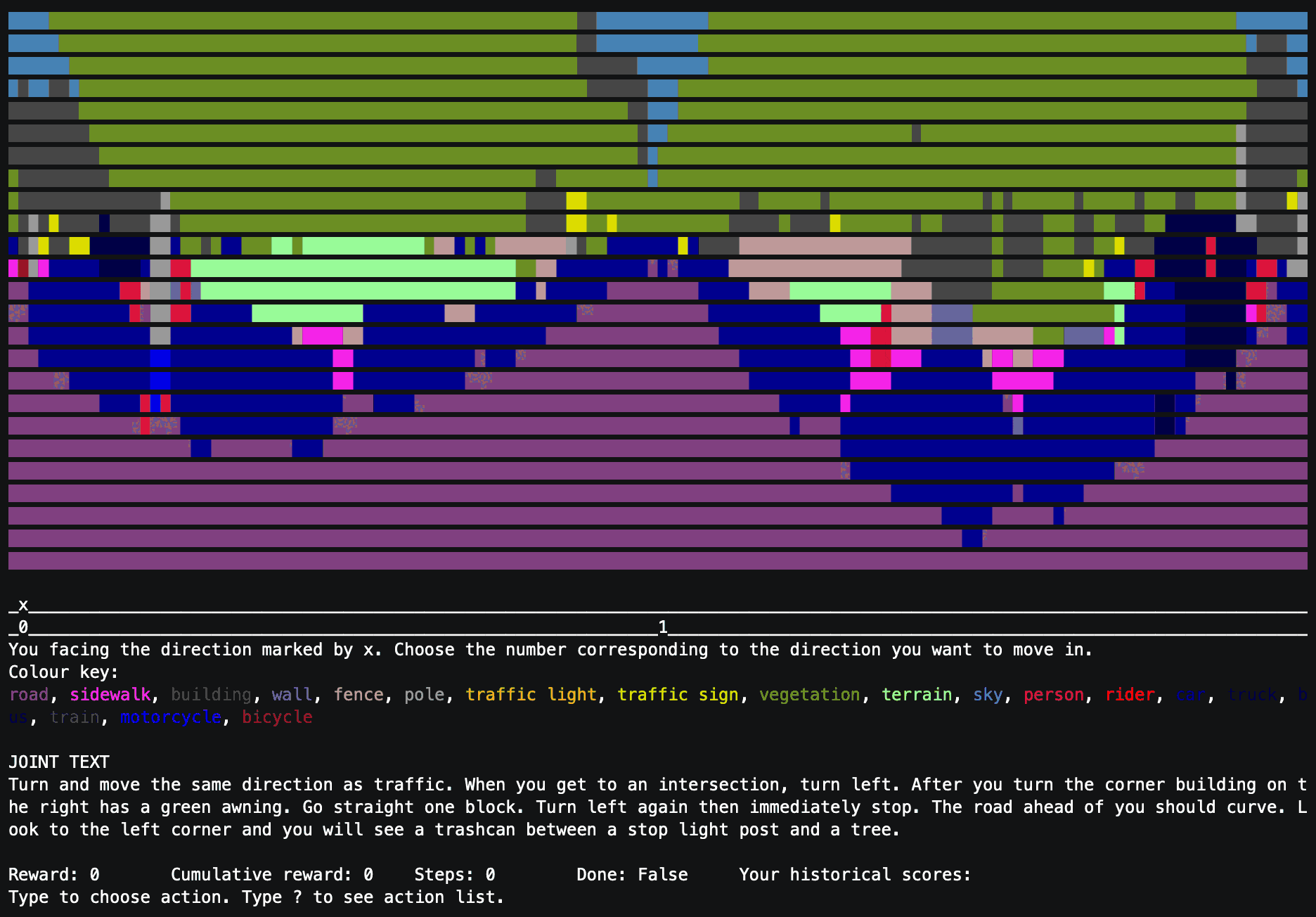}
    \vspace{-0.05in}
    \caption{The~\benchmarkname~\texttt{play} utility (shown here for SymTD) enables human playthrough as well as visualizing what input the model observes. Because all environments are symbolic, the~\texttt{play}utility works in console (e.g.~via~\texttt{ssh},~\texttt{tmux}) without need for X-forwarding.}
    \label{fig:play_interface}
    \vspace{-0.25in}
\end{figure}

The goal of~\benchmarkname~is to provide a simple-to-use benchmark that allows researchers to quickly evaluate methods across all of these environments as well as their respective challenges.
We thus combine these environments under a unified interface built on top of OpenAI Gym~\citep{brockman2016gym}.
In each environment instance, the agent observes text inputs as well as world observations.
For grid worlds such as RTFM, Messenger, and SILGNetHack, the agent receives a 2-D bird's-eye-view symbolic grid as observations.
For visually inspired environments such as ALFWorld and SymTD, the agent receives a symbolic egocentric view of the present scene.
Figure~\ref{fig:play_interface}~shows how~\benchmarkname~environments are rendered to players via the~\texttt{play}~utility.
In the rest of this section, we describe each~\benchmarkname~environment in detail.
Appendix~\ref{app:code}~shows how to use~\benchmarkname~in Python.
Appendix~\ref{app:license}~shows licensing for~\benchmarkname~environments.
\vspace{-0.1in}
%
\paragraph{Selection criteria} We select interactive environments that span the challenges presented in Table~\ref{tab:silg_stats}, \camrdy{are easily converted to symbolic representations, and avoid the use of additional simulators (e.g.~Matterport3D \citep{mattersim}).
While visual perception is clearly important for language grounding~\citep{ferraro-etal2015survey}, we focus on the unique challenges of symbolic environments such as multi-hop reasoning and generalization to rich sets of procedurally generated dynamics.
We leave the challenge of developing a visually rich multi-environment grounding benchmark to future work. 
}
Due to the lack of gold trajectories in many of the selected environments, we do not support imitation learning (IL) in this version of~\benchmarkname.
\vspace{-0.1in}
\paragraph{RTFM}
RTFM~\citep{zhong2020rtfm} is a grid-world environment where an agent interprets text to acquire the correct items to fight the correct monsters.
A key challenge in RTFM is multi-modal multi-step reasoning (at least 6 steps) combining world observations with texts associated with multiple entities.
Given a team to beat, the agent must identify which monster is on the team, then identify the item descriptor that would beat the monster descriptor.
Finally, the agent must acquire the item with the correct descriptor and engage the correct monster to win.
RTFM evaluation is on games with unseen rules,
forcing agents to make novel reasoning steps to generalize successfully.
At each step, the agent receives a symbolic grid containing names of entities present, as well as texts indicating the high level rules, the agent inventory, and the goal of the particular game instance.
We include all 4 RTFM curriculum stages, but only show results for the first stage in this preliminary study.
\vspace{-0.1in}
\paragraph{Messenger}
Messenger~\citep{hanjie21grounding}, is a grid environment where the agent must acquire a message and deliver it to the goal while avoiding an enemy after extracting entity-role assignments from a text manual.
A key challenge in Messenger is the adversarial train-evaluation split without prior entity-text grounding.
There is no overlap in entity-role assignments between training and evaluation, forcing agents to make compositional entity-role generalizations.
At each step, the agent receives a symbolic grid containing symbol IDs of entities present, as well as texts indicating roles of each entity.
The entities are referred to in text by many names, which have no lexical overlap with their symbol ID. \camrdy{That is, the text ``dog'' in the text for example is the non-textual symbol 2 in the observation and the association between entities and references must be learned via interaction.}
We include all 3 Messenger curriculum stages, but only show results for the first stage in this preliminary study.
\vspace{-0.1in}
\paragraph{SILGNetHack}
NetHack is a a complex rogue-like game from the NetHack learning environment~\citep{kuttler2020nethack}.
In SILGNethack, we combine 3 tasks (\texttt{Score},~\texttt{Gold}, and~\texttt{Scout}) and specify the task to complete for each episode via a text prompt.
SILGNetHack is challenging due to its large state space and partial observability.
The agent may descend multiple floors and sections of each floor may be obscured until exploration by the agent.
Because of the different score distributions of each task, we mark a trajectory as successful if it exceeds a task-specific score threshold determined from human playthroughs.
We evaluate agents on previously unseen map layouts that are procedurally generated with new seeds disjoint from the ones used during training.
More information about the~\benchmarkname~multi-task SILGNetHack is in Appendix~\ref{app:nethack}.
At each step, the agent receives a symbolic grid containing symbol IDs of entities present, as well texts denoting the goal, agent stats, and feedback from the environment after the agent's last action.
SILGNetHack vocabulary is technically infinite because players can arbitrarily name things, however in our expert playthroughs of~\benchmarkname~SILGNetHack, we observe just over 100 unique words.
Human experts win just under 80\% of games with an average of 34 steps, which demonstrates the challenge of SILGNetHack.
All failures can be attributed to hitting the step limit before acquiring the necessary win conditions.
\vspace{-0.1in}
\paragraph{ALFWORLD (text ALFRED)}
In ALFWorld, an agent navigates and manipulates objects inside a 3D kitchen~\citep{shridhar2021alfworld}.
Its large text action space, with more than 50 valid actions \camrdy{(given by the game engine)} for most scenes is a key challenge.
Unlike its visual counterpart ALFRED~\citep{shridhar2020alfred} where the agent observes 3-D images of the kitchen, in ALFWorld the agent must rely on language descriptions of the kitchen.
Goals are provided in human written language (e.g.~put a clean sponge on the metal rack).
The language in ALFWORLD is not complex, but are 100 words on average due to a large number of items in a single scene.
Following recent work~\citep{shridhar2021alfworld}, we evaluate on both unseen instructions (new instr) and unseen room layouts (new layouts).
At each step, the agent receives the goal text and a list of items present in the room \camrdy{(e.g. ``cup 1'', ``bottle 2'')}. We concatenate the names of these items into a symbolic world observation grid \camrdy{, each entry containing the name of one item}.
The agent then selects from plausible commands given what is present in the scene.
\vspace{-0.1in}
\paragraph{SILGTouchdown (SymTD, VisTD)}
In Touchdown, the agent navigates through Google Street View panoramas according to long compositional instructions that tests spatial reasoning~\citep{chen2018touchdown, mirowski2019streetlearn,mehta2020retouchdown}.
A key challenge is the rich human-written navigation instructions that describe photorealistic images.
Touchdown's long human-written instructions contain many intra-text reference hops, which we approximate as the number of sentences plus the number of sequential connectors such as ``then''.
We convert Touchdown to a symbolic environment by segmentating its panoramas into semantic grids.
In each step, the agent observes the instruction text and a grid of discretized segmentation class IDs corresponding to the current panorama.
It then chooses among a list of radial directions to proceed to the next panorama.
The agent wins if it passes the goal location.
We use the same train-test split as the original Touchdown environment, which features unseen navigation texts.

We show that our symbolic Touchdown (SymTD) facilitates faster learning compared to learning in its visual equivalent (VisTD).
Human performance demonstrates some limitations of SymTD, with an expert win rate just over 60\%.
This may be due to the symbolic representations removing information referenced by the instructions such as color, or because the segmented features are visually disparate from real-world views \citep{dubeyICMl18humanRL}.
We also include manual stop variants of SymTD and VisTD, which are functionally equivalent to the original Touchdown.
Appendix~\ref{app:touchdown}~details these variants,  SymTD/VisTD creation \camrdy{as well as discussions on human performance}.
Compared to prior work on Touchdown and ALFWorld, we train using RL without supervised trajectories as opposed to imitation learning.

\section{The~\modelname~Baseline Model}

\begin{figure}[t]
    \centering
    \includegraphics[width=\linewidth]{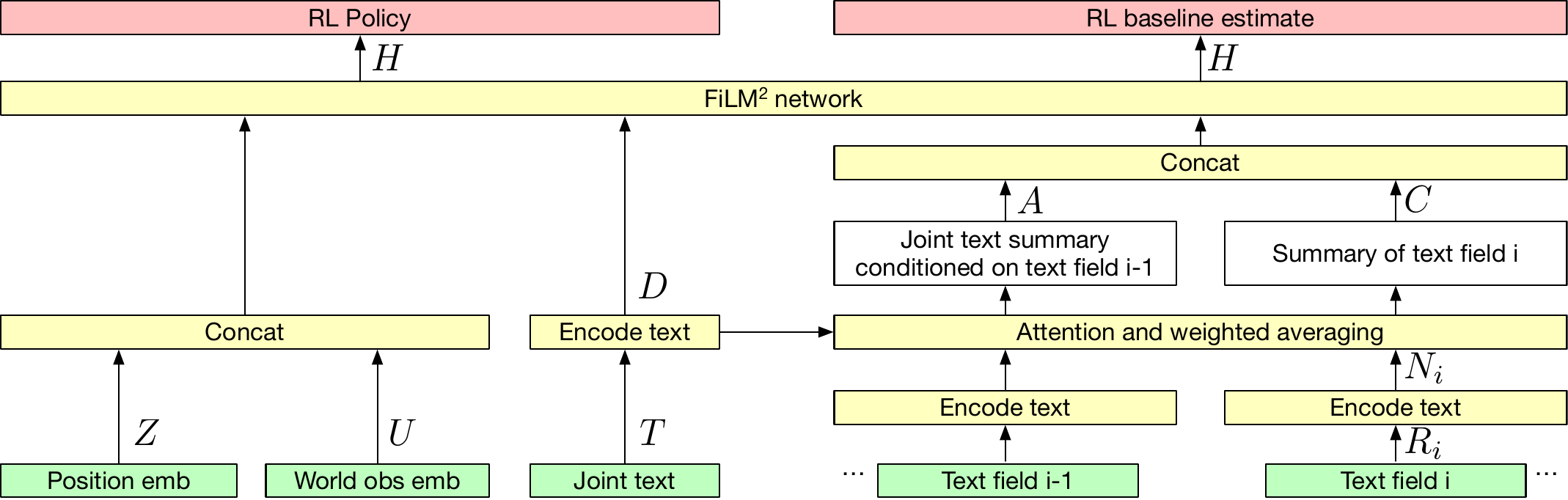}
    \caption{The~\modelname~(\modelnameshort) baseline. Inputs are green, intermediate results white, outputs red, and model components yellow. Details about the~\film~layer is in Appendix~\ref{app:film}.}
    \label{fig:baseline}
    \vspace{-0.2in}
\end{figure}

Figure~\ref{fig:baseline} shows the~\modelnameshort~baseline for the~\benchmarkname~benchmark.
To the best of our knowledge, this is the first shared model architecture for RTFM, Messenger, NetHack, ALFWorld, and Touchdown.
Consider an agent situated in an arbitrary~\benchmarkname~environment.
At each time step $t$, the model receives from the environment the following inputs (precise inputs for each environments are shown in Appendix~\ref{app:experts}).
\begin{itemize}[labelindent=0pt]
    \item \textbf{World observations} $\world \in \real{\worldheight \times \worldwidth \times \lent}$ where $\worldheight$ and $\worldwidth$ are the height and width of the observation and each element corresponds to the $\lent$-word symbol ID(s) of its content.
    \item \textbf{Joint text} $\dynamics \in \real{\ldynamics}$ of $\ldynamics$ tokens of the text to attend over.
    \item \textbf{Text fields} $\nondynamics \in \real{\nnondynamics \times \lnondynamics}$ where the $i$th row contains the $i$th of $\nnondynamics$ environment text field such as agent inventory or environment feedback. $\lnondynamics$ is the max token count of these texts.
    \item \textbf{Relative position} $\relpos \in \real{\worldheight \times \worldwidth \times 2}$ \camrdy{cell-wise feature that denotes the position of each cell relative to the player agent in the $x$ and $y$ directions.}
\end{itemize}
As a policy learner, the model must output a distribution $Y$ over the action space.
We additionally output a baseline estimate of the value function to stabilize policy learning~\citep{espeholt18a}.
Let $\demb$ and $\drnn$ denote embedding and bidirectional LSTM sizes.
We first sum embeddings for each cell in the world observation to obtain world representation $\repworld = \fsum{\embed{\world}} \in \real{h \times w \times \demb}$.
Next, we encode the $i$th text field $\nondynamics_i$ and the joint text $\dynamics$ using an bidirectional LSTMs~\citep{hochreiter1997long}.
\begin{eqnarray}
\repnondynamics_i &=& \BiLSTM{\repnondynamics}{\embed{\nondynamics_i}} \in \real{\lnondynamics \times \drnn} \\
\repdynamics &=& \BiLSTM{\repdynamics}{\embed{\dynamics}} \in \real{\ldynamics \times \drnn}
\end{eqnarray}
We then compute weighted average over text fields $\tilde{\sanondynamics}_i$ and attention $\tilde{\attnnondynamics}_i$ over the joint text.
\vspace{-0.05in}
\begin{eqnarray}
    \tilde{\sanondynamics}_i &=& \selfattn{i}{\repnondynamics_i} = \sum_j \softmax{\linear{i}{\repnondynamics_i}}_j \repnondynamics_{ij} \in \real{\drnn} \\
    \tilde{\attnnondynamics}_i &=& \attend{\repdynamics, \tilde{\sanondynamics}_i} = \sum_j \softmax{ \repdynamics \tilde{\sanondynamics}_i}_j \repdynamics_j \in \real{\drnn}
\end{eqnarray}
\vspace{-0.05in}
We compress $\tilde{\sanondynamics} \in \real{\nnondynamics \times \drnn}$ and $\tilde{\attnnondynamics} \in \real{\nnondynamics \times \drnn}$ again to support any number of text fields.
 
\begin{tabular*}{\textwidth}{ll}
\begin{minipeqn}
  \sanondynamics = \selfattn{\sanondynamics}{\tilde{\sanondynamics}} \in \real{\drnn}
\end{minipeqn}&
\begin{minipeqn}[b]
  \attnnondynamics = \selfattn{\attnnondynamics}{\tilde{\attnnondynamics}} \in \real{\drnn}
\end{minipeqn}
\end{tabular*}
We now have representations for world observations $\repworld$, text fields $\sanondynamics$, and joint text conditioned on text fields $\attnnondynamics$.
We apply successive \film~layers to build multiple levels of codependent representations between texts and world observations to model multiple cross-modal reasoning steps~\citep{zhong2020rtfm}.
To support arbitrary number of text fields, we modify the text input of the $i$th \film~layer to be the concatenation of the text fields $\sanondynamics$, attention over joint text conditioned on text fields $\attnnondynamics$, and attention over joint text conditioned on the visual summary of the last \film~layer $s^{(i-1)}$.
\begin{eqnarray}
V^{(i)}, s^{(i)} & = & \mathrm{FiLM}^2 \left( \concat{V^{(i-1)}; \relpos}, \concat{\sanondynamics, \attnnondynamics, \attend{\repdynamics, s^{(i-1)}}} \right)
\end{eqnarray}
We use the definition of \film$\left({\mathrm{visuals}}, {\mathrm{texts}}\right)$ from~\citet{zhong2020rtfm} and summarize its intuition and computation in Appendix~\ref{app:film}.
We define $V^{(1)}$ and $s^{(1)}$ to be the initial world observation $\repworld$ and its spacial max-pooling.
Finally, we use a multi-layer perceptron to build a fixed-size codependent representation of the inputs based on the last \film~layer's output $\repfinal = \tanh{\left(\linear{4}{\flatten(V^{(\mathrm{last})}}\right)}$, which is used to compute the baseline estimate of the value function $\baseline = \mlp{\baseline}{\repfinal}$ and the policy $Y(\repfinal)$ expressed as a distribution over actions.
%
%
While the core architecture of~\modelnameshort~is identical for all environments, a different policy module $\policy$ is necessary for different types of action spaces.
\vspace{-0.5em}
\paragraph{Fixed sized action space (RTFM, Messenger, SILGNetHack)}
We simply apply a multilayer perceptron to the final representation $\policy = \mlp{\policy}{\repfinal}$.
\vspace{-0.1in}
\paragraph{Multiple-choice text action space (ALFWorld)}
Let $\cmd_j$ denote tokens for the $j$th choice (e.g.~pick up the mug), which we encode a bidrectional LSTM $\repcmd_j = \BiLSTM{\repcmd}{\embed{\cmd_j}}$.
We then attend over this text using the final representation $\repfinal$ to score for $j$th choice $\policy_j = \linear{4}{\attend{\repcmd_j, \repfinal}}$.
\vspace{-0.2in}
\paragraph{Multiple-choice navigation action space (SILGTouchdown)}
Let $j$ denote the index of the world representation corresponding to a movement direction.
For example, for a world observation width of 100, the index corresponding to advancing in the 30 degrees direction is $\frac{30 * 100}{360} \approx 8$.
We encode the navigation choice by selecting its corresponding world observation representation, then scoring it via dot product with the final output representation $\policy_j = \linear{5}{\repworld_{j}} ^ \intercal \repfinal$.

\section{Experiments}

\paragraph{Setup}
How well does a shared architecture do across all five~\benchmarkname~environments?
To answer this, we train and evaluate~\modelnameshort~using Torchbeast~\citep{torchbeast2019}, a distributed RL framework with importance weighted actor-learners based on IMPALA~\citep{espeholt18a}.
For each environment (separately), we train on training, do early stop on validation, and evaluate on test.
NetHack does not distinguish between train and evaluation, hence we create our own splits by dividing the seed range (first 1 million seeds for training, second for validation, and third for test).
We run 5 random seeds for each environment.
The hyperparameter and compute resources are respectively shown in Appendix~\ref{app:hyperparam} and~\ref{app:compute}.
\benchmarkname.

\begin{figure}[t]
    \centering
    \includegraphics[width=\linewidth]{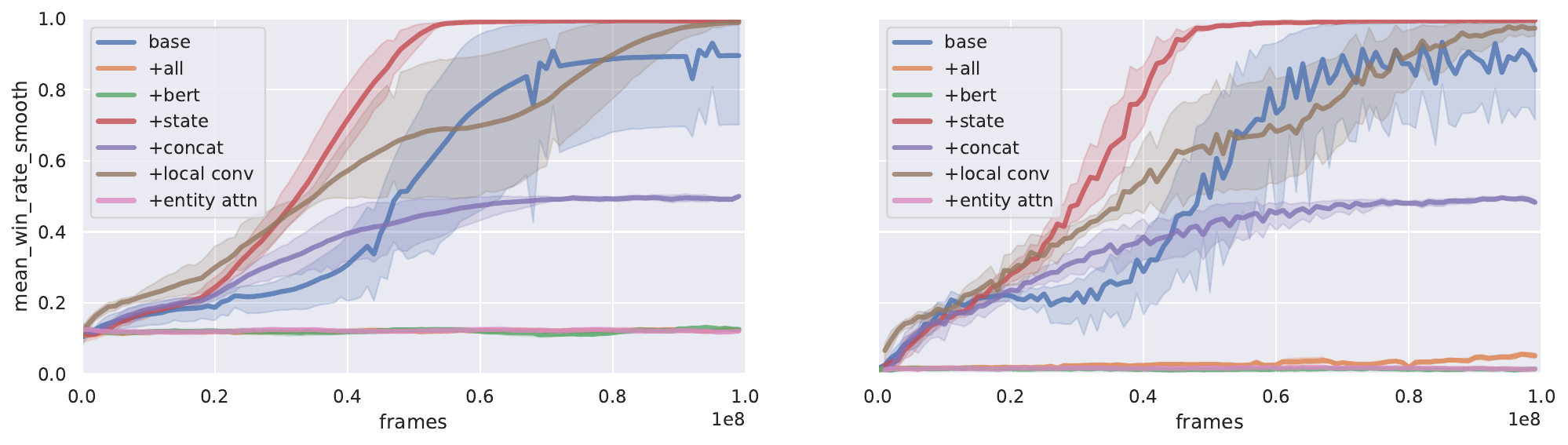}
    \vspace{-0.2in}
    \caption{RTFM performance. Left: train envs, right: validation envs.}
    \label{fig:performance_rtfm}
    \vspace{-0.1in}
\end{figure}

\begin{figure}[t]
    \centering
    \includegraphics[width=\linewidth]{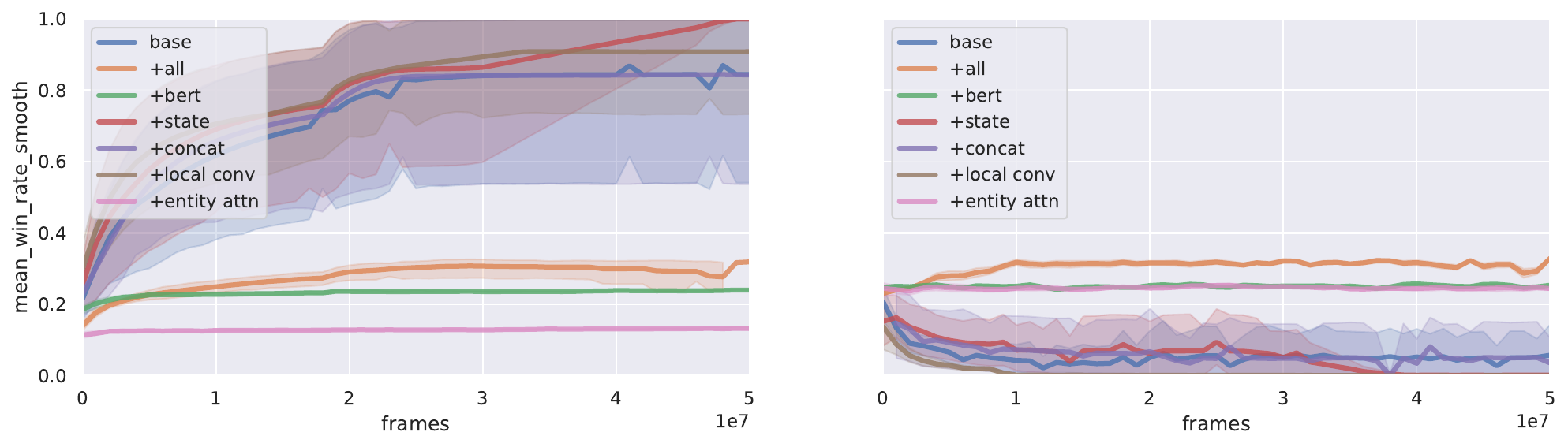}
    \vspace{-0.2in}
    \caption{Messenger performance. Left: train envs, right: validation envs.}
    \label{fig:performance_messenger}
    \vspace{-0.1in}
\end{figure}

\begin{figure}[t]
    \centering
    \includegraphics[width=\linewidth]{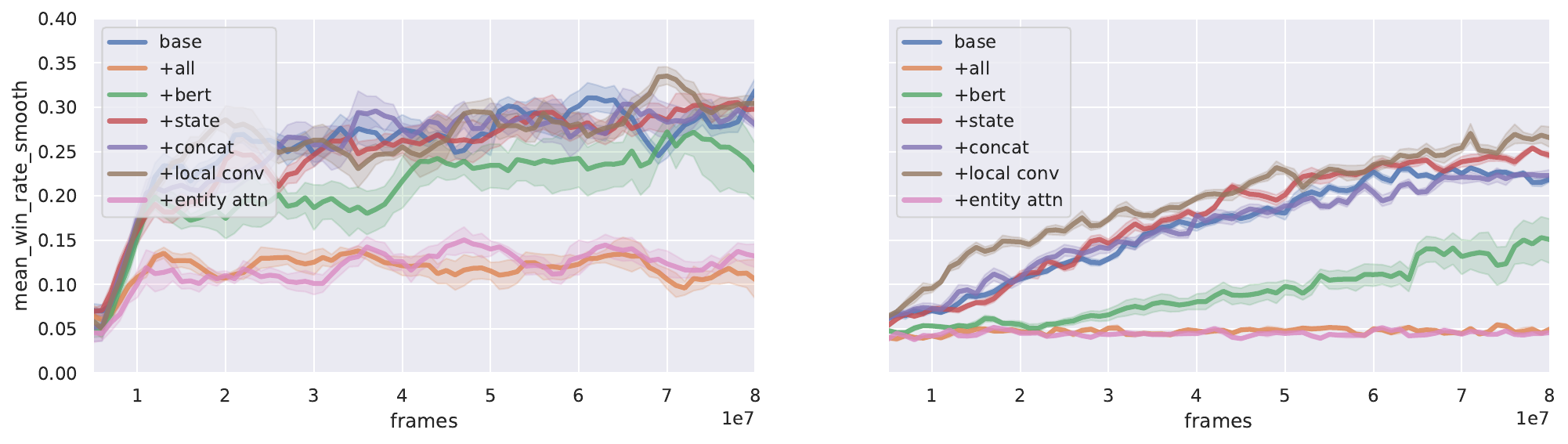}
    \vspace{-0.2in}
    \caption{SILGNetHack performance. Left: train envs, right: validation envs.}
    \label{fig:performance_nethack}
    \vspace{-0.1in}
\end{figure}

\begin{figure}[t]
\centering
  \includegraphics[width=\linewidth]{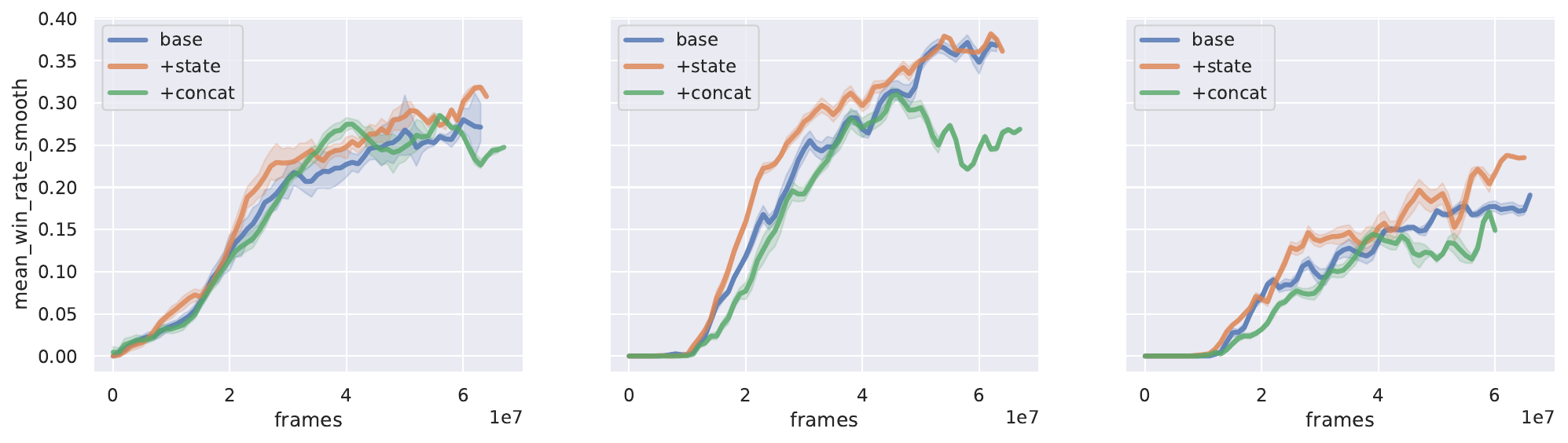}%
\caption{ALFWorld perfomance. Left: train envs, middle: new instruction validation envs, right: new instruction+new layouts validation envs. For efficiency we only evaluate on a subset (50 out of 140) of the validation environments for early stopping. We do train BERT variants here due to computational constraints. ALFWorld does not have entity IDs and no agent location, hence we do not show local convolution nor entity attention experiments.}
\label{fig:performance_alfworld}
\vspace{-0.1in}
\end{figure}

\begin{figure}[t]
    \includegraphics[width=\linewidth]{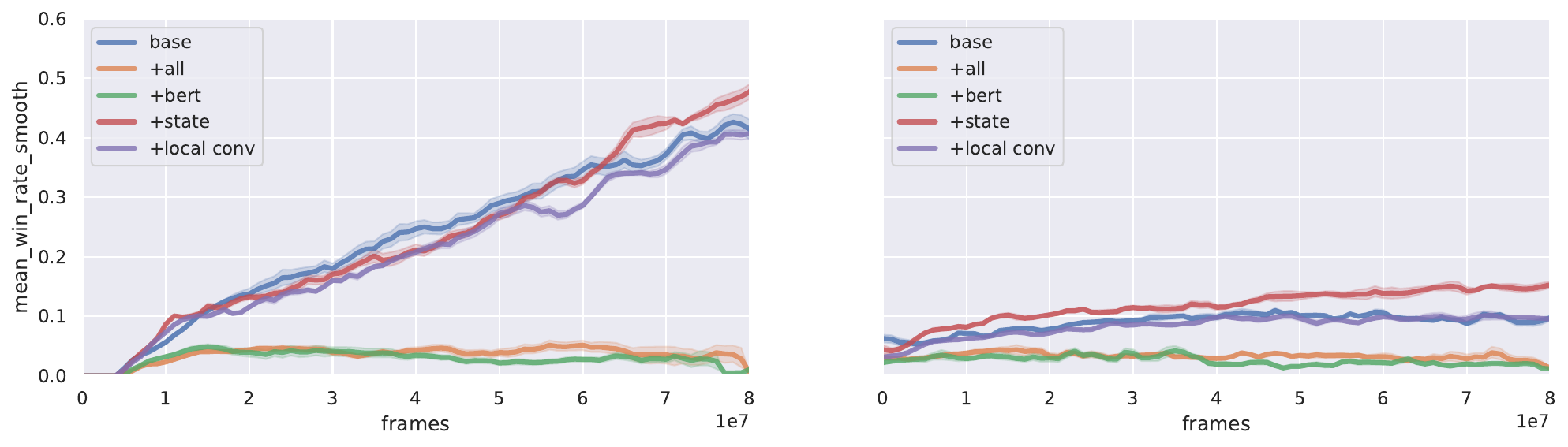}
    \vspace{-0.25in}
    \caption{SymTD performance. Left: train envs, right: validation envs. Touchdown does not have entities, hence we do not show experiments for entity attention.}
    \label{fig:performance_td}
    \vspace{-0.15in}
\end{figure}

\vspace{-0.1in}
\paragraph{Results}
Figures~\ref{fig:performance_rtfm} through~\ref{fig:performance_td} show learning curves for each environment.
Table~\ref{tab:test}~shows the test performance for the baseline model and the best model variant.
Despite sharing the same core model architecture,~\modelnameshort~achieves reasonable performance across all environments except Messenger, where it overfits due to lack of pretrained LM and entity-centric attention.
Nevertheless, the best performing model significantly trails human performance, indicating room for further improvement.

\begin{table}[t]
\centering
\caption{Success rate on test environments for~\modelnameshort~and its best variant. Standard deviation are in brackets. We early stop on validation and evaluate best checkpoint on test.
For RTFM, Messenger, and SILGNetHack, we evaluate 100 episodes.
For ALFWorld and Touchdown, we evaluate on initial states from each test episode.
\camrdy{The variant with best performance across envs is {\tt +state}.}
The SOTA for RTFM, Messenger, and ALFWorld are respectively from~\citet{zhong2020rtfm},~\citet{hanjie21grounding}, and~\citet{shridhar2021alfworld} (\camrdy{std was not reported in ALFWorld}).
\textsuperscript{$\vartriangle$}SOTA for ALFWorld relies on supervised trajectories and beam search, which~\modelnameshort~does not use.
There are no previous results for multitask SILGNetHack and SymTD as they are introduced here.
Though not comparable, the manual stop VisTD SOTA trained using imitation learning on supervised trajectories is 16.7\%~\citep{xiang2020learningToStop}.
}
\label{tab:test}
\begin{tabular}{@{}lllllll@{}}
\toprule
Model & RTFM & Messenger & SILGNetHack & \multicolumn{2}{c}{ALFWorld} & SymTD \\ \cmidrule(lr){5-6}
 &  &  &  & new inst & new inst+layouts &  \\ \midrule
Base & 88.8 (22.4) & 0 (0) & 23.8 (0.8) & 21.0 (1.5) & 16.0 (2.1) & 9.7 (1.3) \\
Best & \texttt{+state} & \texttt{+all} & \texttt{+local conv} & \texttt{+state} & \texttt{+state} & \texttt{+state} \\
 & 99.2 (0.7) & 31 (2.6) & 25.4 (3.3) & 23.6 (2.8) & 16.6 (2.9) & 14.9 (1.8) \\ \midrule
SOTA & 83 (21) & 85 (1.4) & N/A & 40\textsuperscript{$\vartriangle$} & 37\textsuperscript{$\vartriangle$} & N/A \\
\midrule
Human & 100 & 100 & 78.1 & 100 & 100 & 61.5 \\ \bottomrule
\end{tabular}
\end{table}


\subsection{Analyses of recent grounded language RL modelling contributions}

Next, we use~\benchmarkname~to evaluate recent modelling advances for language grounding across environments by adding them to the~\modelnameshort~baseline. \camrdy{These modelling enhancements were proposed for (and resulted in key gains on) the environments included in}~\benchmarkname.
Namely, we analyze the effectiveness of recurrent state-tracking, entity-centric local convolution, entity-centric attention, and pretrained LMs.

\vspace{-0.1in}
\paragraph{Recurrent state tracking (\texttt{state})}
As in~\citet{kuttler2020nethack}, we augment the~\modelnameshort~baseline with a state-tracking LSTM by replacing the final $\repfinal$ with $\repfinal^\prime = \repfinal + \LSTM{\repfinal, \lstmstate_{t-1}}$, where $\lstmstate_{t-1}$ is the previous LSTM state (summing LSTM output and $\repfinal$ outperforms replacing $\repfinal$ with LSTM output).
State-tracking consistently improves convergence and generalization, even when the correct next step is fully determined by current world observations (e.g.~RTFM).
This may be because it helps prevent local minima that cause repetitive actions.
The exception to this is Messenger, where state-tracking does not help generalize to the evaluation distribution.
\vspace{-0.1in}
\paragraph{Entity-centric local convolution (\texttt{local conv})}
\citet{hill2019environmental}~proposed local convolution around the agent to obtain an egocentric view of world observations.
While this helps generalize in SILGNetHack, it does not help significantly in other environments.
One reason is that this provides redundant information as positional embeddings, which is already included in the base model and is a cheaper alternative to adding an additional egocentric convnet.
\vspace{-0.1in}
\paragraph{Entity-centric attention (\texttt{entity attn})}
\citet{hanjie21grounding}~propose replacing entity representations with attention over text specification, such that the world observations are forcibly composed using text representations.
We add this by replacing world representation $\repworld$ with entity attention over text fields $\nondynamics$ as described in~\citet{hanjie21grounding}.
This constraint causes underfitting of SIR on most environments.
\camrdy{Since the entity representation is built entirely using the text, when there is incomplete entity information or it is difficult to extract the relevant information from the manual text this can be a handicap.}
However, for Messenger, entity-centric attention prevents overfitting.
\vspace{-0.1in}
\paragraph{Pretrained language model (\texttt{bert})}
A natural question in language-grounding is how to leverage large, pretrained LMs~\citep{hill2020human}.
We use a simple method to incorporate BERT~\citep{devlin2018bert} by replacing all text encoding with the summation of the original bidirectional LSTM encoding and BERT encoding.
Due to the memory requirement of large pretrained LMs, we cannot fine-tune the LM during training, and thus keep the LM parameters fixed.
Pretrained LMs (\texttt{bert} and \texttt{all}) help generalization in Messenger but does not improve performance on other environment in our experiments.
\camrdy{For tasks such as RTFM and SILGNetHack, our use of a general-purpose LM may not be beneficial for the highly specific language used in those tasks (i.e.~fantasy world with word like shaman, goblin, mage etc). We stress that this is a preliminary investigation into the use of LMs on these environments, and we encourage future research on how to effectively use pretrained LMs across environments using~\benchmarkname.}
\subsection{Analyses of~\benchmarkname~environments}
\vspace{-0.1in}
Finally, we examine performance of~\modelnameshort~and variants to analyze  challenges presented by~\benchmarkname.
\vspace{-0.1in}
\paragraph{Generalization requirement of environments}
\benchmarkname's evaluation environments require different types of generalization.
RTFM requires generalizing to new environment dynamics by referring between world observations and multiple texts;
because~\modelnameshort~adopts~\film~from~\citet{zhong2020rtfm}, it is able to achieve such generalization. \camrdy{Messenger requires compositional entity-role generalizations. That is, if an entity (e.g.~dog) has a certain role (e.g.~message holder) in training, such an entity-role assignment never appears in validation or test.}
~\modelnameshort~quickly overfits to to entity-role assumptions  (e.g. dog as the message) in training \camrdy{suggesting the need for additional work on achieving this type of generalization using a joint model architecture}.
Combining pretrained LM with other enhancements (\texttt{+all}) results in generalization improvement, however the convergence remains very slow.
This suggests that generalizing to new dynamics across environments without obvious lexical cues from the text remains a difficult challenge.
SILGNetHack and ALFWorld require generalizing to new procedurally generated scenes, which~\modelnameshort~achieves.
In the additional out-of-domain ALFWorld evaluation where the model must generalize to new layouts, state-tracking allows the model to generalize faster.
Touchdown requires generalizing to new natural language instructions.
Here, the baseline suffers from a large generalization gap.
We hypothesize that more effective means of incorporating pretrained LMs is necessary to achieve this type of generalization.
\vspace{-0.1in}
\paragraph{Necessity of separate text fields}
In \texttt{concat}, we concatenate text fields into a single string, which we encode using a bidirectional LSTM.
In this case, both joint text $\repdynamics$ and text field representations $\repnondynamics$ are set to this encoding.
This degrades performance especially in RTFM, which shows that multi-hop references is more easily learned when the text fields are separated and modeled via structured attention.
Note that this model variant is not shown for Touchdown because it only has one text field.

\begin{table}[t]
\centering
\caption{Transfer task success rate from symbolic to visual envs for baseline and its best variant. Standard deviation shown in brackets. For ALFRED, we give ALFWorld-trained models language templates filled with detected objects from vision using oracle and Masked-RCNN object detectors.}
\label{tab:vis_transfer}
\begin{tabular}{@{}lllllllcc@{}}
\toprule
Model & \multicolumn{6}{c}{ALFWorld/ALFRED} & SymTD & VisTD \\ \cmidrule(lr){2-7}
 & \multicolumn{3}{c}{new inst} & \multicolumn{3}{c}{new inst + layouts} & to  &  \\ \cmidrule(lr){2-7}
 & text & oracle & m-rcnn & text & oracle & m-rcnn & VisTD &  \\ \midrule
Base & 21.0(1.5) & 11.2(3.3) & 3.0(1.7) & 16.0(2.1) & 0.7(0.4) & 0.3(0.2) & 9.7(1.3) & 4.3(4.0) \\
Best & \texttt{+state} & & & \texttt{+state} & & & \texttt{+state} & \texttt{base} \\
 & 23.6(2.8) & 11.3(1.9) & 7.1(1.1) & 16.6(2.9) & 1.3(1.1) & 0.7(0.6) & 14.9(1.8) & 4.3(4.0) \\ \bottomrule
\end{tabular}
\vspace{-0.1in}
\end{table}

\vspace{-0.1in}
\paragraph{Learning from symbolic vs visual world observations}
Table~\ref{tab:vis_transfer} shows that policies learned in the symbolic environment transfer to the 3-D environment.
Using oracle and Masked-RCNN~\citep{he2017maskrcnn}, the ALFWorld policy can be transferred by filling observation text templates using detected objects.
Our result with oracle detector is in line with~\citet{shridhar2021alfworld}, though our performance is weaker because we do not use annotated data nor DAgger~\citep{ross2011dagger}.
As with prior results, transfer to visual worlds with new layouts remains very challenging~\citep{shridhar2021alfworld}.
Transfer using Masked-RCNN results in large drop in performance, nevertheless~\benchmarkname~allows perception, albeit an important challenge, to be factored out so that one can focus and quickly iterate on abstraction challenges.
Table~\ref{tab:vis_transfer}~also shows that models trained on SymTD outperform those trained on VisTD (where $\repworld$ is 10-dim PCA features from a ResNet~\citep{He2015resnet} panorama encoding) despite being faster (383 for SymTD vs. 344 frames per second for VisTD).
That is, by applying segmentation to obtain SymTD, we are able to obtain a better policy than training directly with visual features using VisTD.
The results from both ALFWorld and SymTD show that learning in faster symbolic environments such as~\benchmarkname~can transfer to their visual counterparts, and allows certain perception challenges to be factored out. 
\vspace{-0.1in}
\paragraph{Future work}
We find that some of the most challenging aspects of situated interactive language grounding include (1) grounding text references to entities without lexical overlap, (2) choosing from large textual action spaces, and (3) interpreting complex natural language descriptions.
On the methodology front, further work is needed to investigate how to effectively use pretrained LMs for language grounding.
Moreover, apart from recurrent state tracking, the other model enhancements do not yield significant gains on environments other than the ones they were proposed for.
These results highlight the need for modelling techniques that generalize across environments.

\camrdy{SIR suggests that with additional improvements, it may be possible to have a performant model with the same architecture (but trained independently) across environments. Future work may explore whether (1) a single model with the same parameters can accomplish all tasks, (2) a single model with pretraining can be quickly finetuned on each task, and (3) learning in one environment is transferable to another.} We believe~\benchmarkname~is well-suited to help answer these questions. Furthermore, ~\benchmarkname \camrdy{ is designed to be easily extensible, with opportunities to add additional environments in the future.}

\vspace{-0.1in}
\section{Related Work}
\vspace{-0.1in}

\paragraph{Benchmarks for NLP and RL.}
NLP benchmarks helped the development of models that generalize across different tasks~\citep{wang-etal-2018-glue,wang2019superglue}.
Similar benchmarks have furthered research in RL~\citep{chevalierboisvert2018minigrid,cobbe2019procgen,tassa2020dmcontrol}.
\benchmarkname~is the first benchmark for symbolic interactive language grounding with a diverse set of language and RL challenges.
\benchmarkname~evaluates generalization to new dynamics with (RTFM) and without lexical cues (Messenger) between text and entities, large partially observed worlds (SILGNetHack), large actions spaces (ALFWorld), and complex natural language instructions in rich visual scenes (SymTD).
Finally,~\benchmarkname~provides a standard interface for symbolic interactive grounding environments via Gym, and considers the transfer to their visual counterparts (ALFWorld, SymTD).
For reference, there is a host of perception-rich embodied environments not included in~\benchmarkname~due to the latter's emphasis on symbolic environments~\citep{MacMahon2006WalkTT,misra2017mapping,Das2018EmbodiedQA,mattersim,Ku2020roomAcrossRoom}.
This emphasis allows~\benchmarkname~to provide an efficient benchmark for situated interactive language grounding.
There are other complementary symbolic language grounding environments not included in this initial release of~\benchmarkname~due to time consideration such as~\citep{chevalierboisvert2018minigrid,Ruis2020ABF,samvelyan2021minihack}.
We look forward to incorporating these in future iterations.

\vspace{-0.1in}
\paragraph{Interactive language grounding}
Language grounded policy-learning has been explored in the context of instruction following in tasks like navigation~\citep{chen2011learning, hermann2017grounded, fried-etal-2018-unified,wang2019reinforced,daniele2017navigational,misra2017mapping,janner2018representation}, games~\citep{golland-etal-2010-game,reckman2010learning,andreas2015alignment,bahdanau2018learning,kuttler2020nethack}, and robotic control~\citep{walter2013learning,hemachandra2014learning,blukis2019learning}.
Touchdown, NetHack, and ALFWorld are three examples of such work included in~\benchmarkname.
While the above environments typically assume a small fixed set of world dynamics, other work explores settings where an agent must read text manuals to formulate appropriate policies for the game at hand.
\citet{branavan2012learning} developed an agent to play Civilization more effectively by reading the game manual. 
\citet{narasimhan2018grounding} and \citet{zhong2020rtfm} used text descriptions of game dynamics to learn policies that generalize to new environments and dynamics, without requiring feature engineering.
Unlike these two works, \citet{hanjie21grounding} does not assume initial lexical overlap between entities in the world and entity references in the text manual.
RTFM and Messenger are two examples of such work included in~\benchmarkname.

\vspace{-0.1in}
\paragraph{Generalization to new environments in interactive language grounding}
In previous instruction following work, evaluation environments typically differ from training in their world observations.
These difference range from differences in object placement in the same/new rooms (e.g.~ALFWorld) to procedural generation of large game levels (e.g.~NetHack).
Moreover, some study generalization to new compositional instructions (e.g.~Touchdown).
Recent works explore generalization to new environment dynamics, which must be inferred by reading.
These range from multi-step reasoning across texts (e.g.~RTFM) to grounding entities to new text references (e.g.~Messenger).
The environments in~\benchmarkname~explore a variety of these generalization challenges.
Many modelling techniques have been proposed to address these generalization challenges, including environmental variations~\citep{hill2019environmental}, memory structures~\citep{hill2020grounded}, pretrained language models~\citep{hill2020human}, incremental guidance~\citep{co2018guiding}, subgoal-specification~\citep{andreas2016modular}, and hierarchical RL~\citep{oh2017zero}.
Our baseline and analyses explores some of these techniques, including bidirectional feature-wise linear modulation~\citep{zhong2020rtfm}, recurrent state-tracking~\citep{kuttler2020nethack}, entity-centric convolution~\citep{kuttler2020nethack}, entity-centric attention~\citep{hanjie21grounding}, and  pretrained language modelling~\citep{hill2020human}.

\vspace{-0.1in}
\section{Conclusion}
\vspace{-0.1in}
We introduced~\benchmarkname, a new benchmark for evaluating language grounded agents across unique challenges posed by five symbolic interactive environments.
Using~\benchmarkname, we proposed the first shared architecture and analyzed recent methodological advancements in grounded language learning across on these environments.
We showed that a shared architecture achieves comparable result to environment-specific methods, and that most advances do not result in significant gains on environments other than the one they were designed for.
This highlights the need for modelling techniques that generalize across environments.
Finally, the most models significantly trail human performance on~\benchmarkname, which suggests ample room for future work.
We hope that~\benchmarkname{} will provide a unified platform for evaluating future methodological advances.

\section*{Acknowledgements}
We are grateful to members of UW NLP, Princeton NLP, and Facebook AI Research for their feedback, as well as the anonymous reviewers for their helpful comments and suggestions.
In particular, we thank Howard Chen for detailed discussion on Touchdown and Shunyu Yao on the manuscript.
Moreover, we thank Yoav Artzi, Jesse Thomason, Edward Grefenstette and Tim Rockt\"{a}schel for their invaluable feedback during the initial stages of this project.
Victor is supported in part by the ARO (AROW911NF-16-1-0121) and by the Apple AI/ML fellowship.
Austin is supported by the Princeton University Graduate Fellowship.

\bibliography{myrefs}
\bibliographystyle{plainnat}

\appendix
\section{Impact statement}
\label{app:impact}
\benchmarkname~facilitates research in reinforcement learning for interactive language grounding.
Real-world applications in this research area range from human-computer interfaces, where users controls a computer interface via natural language specifications, to robotics control, where a robot carries out instructions given by users.
Some positive impact research in this area has to do with accessibility.
For example, such interfaces can allow non-experts or people to use complex software and allow people who are physically unable to operate heavy machinery to do so.

Some potential negative impact this research may have is the lack of interpretability that results from complex policies.
This work uses RL, a general solution that can learn from environmental rewards without annotated data.
This type of learning may result in unintuitive policies that achieve the object in surprising ways (e.g.~a robot that knocks a bowl off the counter while bringing the user a cup of coffee).
Language grounded policy learning, which~\benchmarkname~facilitates, is one way of dictating the direction of policy learning.
However, more research is needed to develop more interpretable and controllable RL techniques.

\camrdy{The language of the individual environments in}~\benchmarkname \camrdy{ are highly specific to a particular setting. For example, NetHack and RTFM are based in the fantasy settings, ALFWorld is in a household setting and Touchdown is in street navigation. This means that the learned grounding on these environments may not generalize to other settings. While}~\benchmarkname \camrdy{ is an initial step, additional work is required to train general purpose agents that can interpret natural language in any setting.}

\section{Using~\benchmarkname}
\label{app:code}

We use OpenAI Gym to create a common interface for all five~\benchmarkname~environments.
For RTFM, Messenger, and ALFWorld, we create wrappers for the original environments such that the output of the Gym environment adheres to the shared interface.
The custom environment variants we create for NetHack and for Touchdown are respective described in detail in Appendix~\ref{app:nethack} and~\ref{app:touchdown}.

To instantiate a~\benchmarkname~environment, the user needs to simply instantiate its Gym instance as follows.

\begin{lstlisting}[language=Python]
from silg import envs
import gym
import random
env = gym.make('silg:td_segs_train-v0', time_penalty=-0.02)
obs = env.reset()
action = random.choice(list(range(len(env.action_space))))  # e.g. 0
obs, reward, done, info = env.step(action)
\end{lstlisting}

The~\texttt{obs}~dictionary then contains the environment outputs specified in Section~\ref{sec:benchmark}.

\section{Bidirectional Feature Wise Linear Modulation layer}
\label{app:film}
Feature-wise linear modulation (FiLM), which modulates visual inputs using representations of text inputs, is an effective method for image captioning~\citep{perez2018film} and instruction following~\citep{bahdanau2018learning}.
\citet{zhong2020rtfm}~extends FiLM to its bidrectional variant~\film, which they show to be effective for modeling joint multi-hop references between visual and multiple text inputs.
We find~\film~to be an effective building block for the tasks considered in~\benchmarkname.

Let + and * symbols denote element-wise addition and multiplication operations that broadcast over spatial dimensions.
Let $\filmfeat_\vtext$ denote a fixed-length $\demb\vtext$-dimensional representation of the text and $\filmmatrixfeat_\vvis$ the representation of visual inputs with height $H$, width $W$, and $\demb_\vvis$ channels.
Let $\conv$ denote a convolution layer.
\film~ first modulates visual features using text features:
\begin{eqnarray}
\vectorgamma_\vtext &=& \weight_\gamma \filmfeat_\vtext + \bias_\gamma \\
\vectorbeta_\vtext &=& \weight_\beta \filmfeat_\vtext + \bias_\beta \\
\filmout_\vvis &=& \relu((1 + \vectorgamma_\vtext) * \conv_\vvis(\filmmatrixfeat_\vvis) + \vectorbeta_\vtext)
\end{eqnarray}
Then, it modulates text features using visual features:
\begin{eqnarray}
\matrixgamma_\vvis &=& \conv_\gamma(\filmmatrixfeat_\vvis) \\ 
\matrixbeta_\vvis &=& \conv_\beta(\filmmatrixfeat_\vvis) \\
\filmout_\vtext &=& \relu((1 + \matrixgamma_\vvis) * (\weight_\vtext \filmfeat_\vtext + \bias_\vtext) + \matrixbeta_\vvis)
\end{eqnarray}
The output of~\film~is the sum of the modulated features $\filmout$ and its max-pooled summary $\filmsumm$ across spatial dimensions.
\begin{eqnarray}
  \filmout &=& \filmout_\vvis + \filmout_\vtext \\
  \filmsumm &=& \maxpool(\filmout)
\end{eqnarray}

\section{Multitask NetHack}
\label{app:nethack}
For NetHack, we create a multi-task environment that uniformly samples between the three tasks~\texttt{Score},~\texttt{Gold}, and~\texttt{Scout}.
Given the sampled task, the agent observes a text string that specifies the goal (e.g.~``get more gold''), in addition to the original environment text feedback to the agent's actions.
For each task, we collect 10 human playthroughs where in a human plays the original NetHack Learning Environment and attempts to get the highest score possible within 50 steps.
The empirical mean of these playthroughs is then used as the task's score threshold.
In the~\benchmarkname~version of multi-task NetHack, the agent receives a reward of 1 if it exceeds the score threshold of the current task, and 0 otherwise.
If the episode terminates without exceeding the score, then the agent receives -1.
We find that this method of reward assignment strikes a balance between the very different reward distributions of the individual tasks (using the raw reward from individual tasks causes the agent to only learn to play~\texttt{Scout}, the dominant task with frequent rewards).
NetHack does not naturally provide train/validation/test splits.
We create our own splits by splitting the seed ranges (1-1,000,000 for train, 1,000,001-2,000,000 for validation, 2,000,001-3,000,000 for test).

\section{SymTD, VisTD, and Touchdown}
\label{app:touchdown}

\begin{figure}[t]
    \centering
    \includegraphics[width=\linewidth]{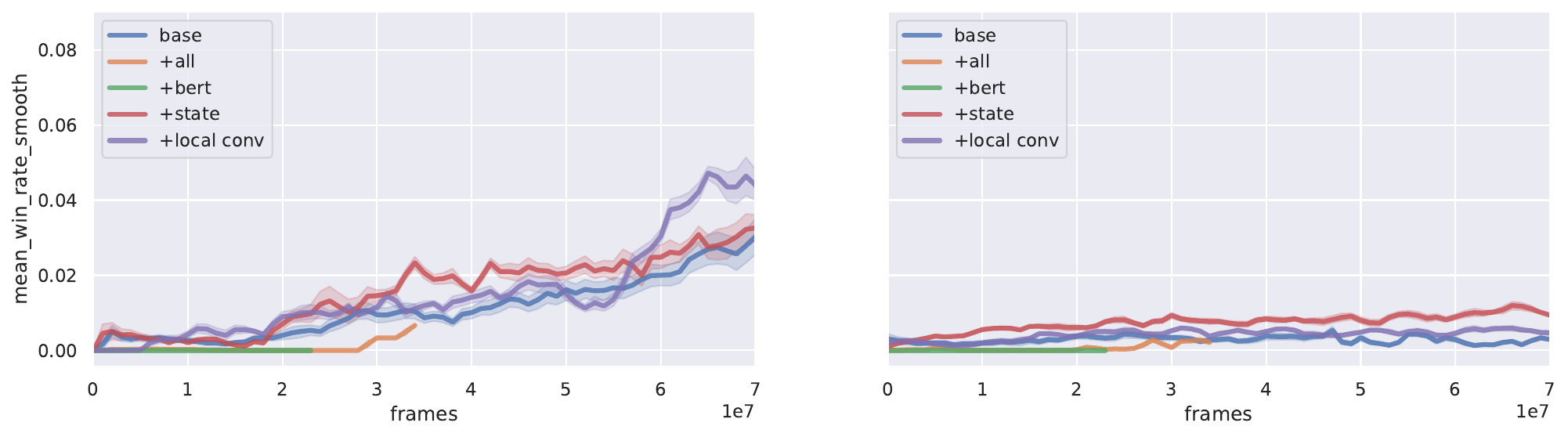}
    \vspace{-0.2in}
    \caption{Manual SymTD performance. Left: train envs, right: validation envs.}
    \label{fig:performance_manual_td_sym}
\end{figure}

\begin{figure}[t]
    \centering
    \includegraphics[width=\linewidth]{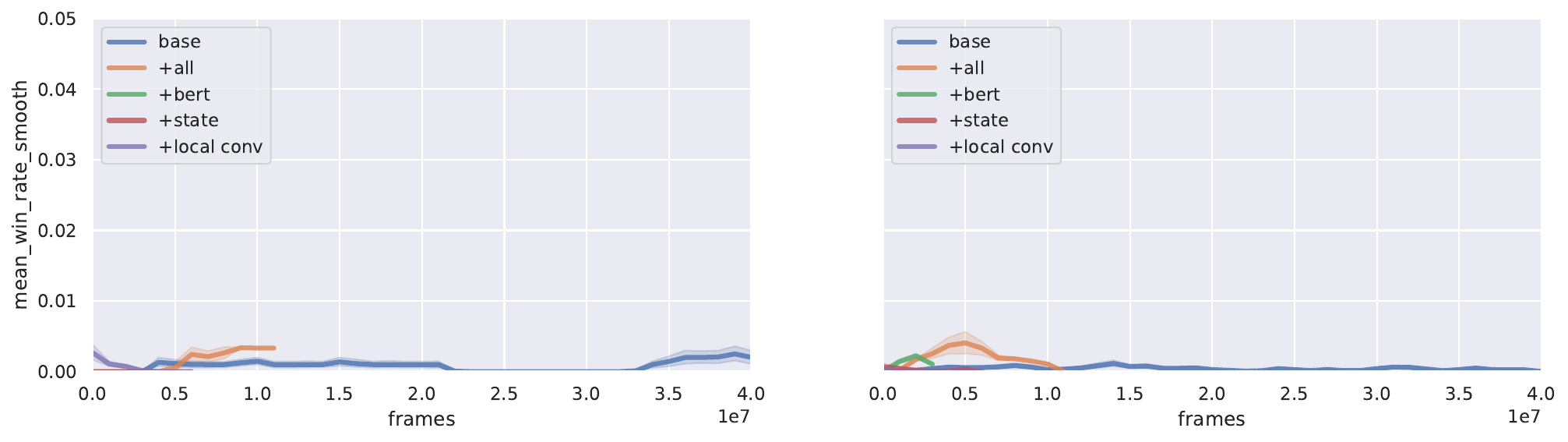}
    \vspace{-0.2in}
    \caption{Manual VisTD performance. Left: train envs, right: validation envs.}
    \label{fig:performance_manual_td_vis}
\end{figure}

\paragraph{Navigation} In the original Touchdown implementation, the agent navigates with left, right, and forward commands.
The left and right commands rotate the panorama at the current node so that the center of the panorama faces an adjacent node.
The forward command then advances the agent to the node currently faced by the agent.
We modify this navigation interface by fixing the panorama and providing the agent with a list of coordinates along the width-dimension of the panorama that corresponds to the locations of adjacent nodes that the agent may advance towards. 
The agent navigates by selecting one of the possible coordinates at each step.
Our implementation allows the agent to see all possible navigation options upfront and reduces trajectory length by eliminating rotations. \camrdy{The setup is similar to}  \cite{fried-etal-2018-unified} \camrdy{except our positional encoding embeds the distance of each point to the agent's current heading along the x-dimension, instead of using angle encodings.}

\paragraph{Rewards} Due to the sparsity of terminal $\pm 1$ rewards, we provide a reward at each step by taking the difference in shortest path graph distance before and after the step (scaled by constant factor). This does not always assign positive reward when following the gold trajectory, but we find that it is a good heuristic in most cases.

\begin{figure}
    \centering
    \includegraphics[width=0.75\linewidth]{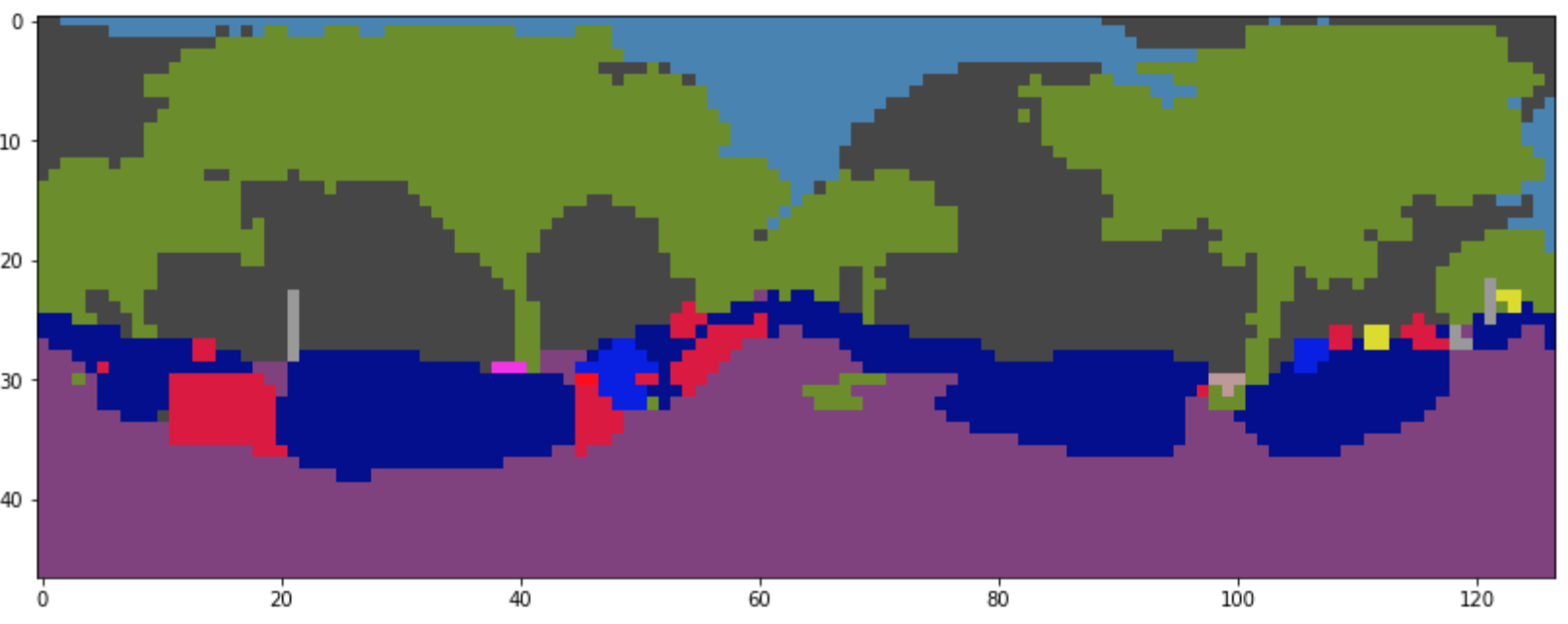}
    \caption{Segmentation map examples to create SymTD. The common objects the colours correspond to are sky (blue), buildings (gray), trees (green), sidewalk (pink), road (purple), cars (blue), traffic lights (yellow), and people (red). Note due to license agreements, this figure is a segmentation done on an example panorama provided directly on the StreetLearn website \url{https://sites.google.com/view/streetlearn/dataset}. Actual segmentations are visually very similar.}
    \label{fig:segmentation}
\end{figure}

\paragraph{SymTD} We pass the raw panoramas from the original Touchdown task through a PSPNet~\citep{zhao2017pspnet} trained on the Cityscapes dataset~\citep{Cordts2016Cityscapes}.
The result is a segmentation map of the raw panorama with identical height and width dimensions.
To allow for caching of the segmentation maps, we downsample the segmentation maps by taking a majority vote in each $23\times23$ patch.
We found that the majority vote caused high-frequency classes (e.g. sky) to drown out low-frequency classes (e.g. pole). Therefore, we scale the vote of each class by its inverse count computed across all segmented panoramas.
If $f(c)$ is the total count for class $c$, and $P$ is a $23\times23$ patch in the segmentation map, the vote for class $c$ in patch $P$ is:
\begin{align*}
    v_P(c) = \frac{1}{f(c)^\alpha}\sum_{p\in P}\mathbbm{1}[p = c]
\end{align*}
The representative class for each patch $P$ is then: $\max_{c\in C}v_P(c)$.
We find that $\alpha = 1$ is effective at generating segmentation maps that preserve low-frequency classes. Figure~\ref{fig:segmentation}~shows examples of such segmentation maps.

\camrdy{We conduct qualitative inspections of a sample of the segmented panoramas and observe that most segmentations are mostly correct relative to the input image. Despite this, human performance remains fairly low at approximately 60\%. The main challenges faced by human players are (1) the symbolic features have no color information and (2) downsampling the segmentations result in highly pixelated figures, such that it is harder to distinguish smaller pedestrians from poles for example and (3) the navigation setup where the current heading is not necessarily the center of the panorama (indicated instead using an x-value) is extremely unintuitive for humans and often leads to the human player becoming disoriented. Given these observations, 60\% may not be the upperbound for SymTD because controls unintuitive to humans do not affect ML models the same way.}

\paragraph{VisTD} We pass the raw panoramas through the ResNet-50~\citep{He2015resnet} backbone of a PSPNet trained on the Cityscapes dataset.
We use the feature map from the last layer.
Due to the large dimensionality along the feature axis and the difficulty caching these for efficient RL, we reduce the number of features using PCA to the top 10 principle components.
The resulting feature maps for each panorama is $47\times128\times10$.

\paragraph{Manual stop TD} In our variant of TD, the agent succeeds and the episode terminates immediately after the agent reaches the target node.
We also include manual variants of SymTD and VisTD where the agent must manually select the "stop" option at the correct node.
Thus, SymTD and VisTD are functionally equivalent to the original Touchdown environment.

The performance of our baseline as well as the baseline with various modelling advances are shown respectively in Figures~\ref{fig:performance_manual_td_sym} and~\ref{fig:performance_manual_td_vis} for Manual SymTD and Manual VisTD.
Compared to SymTD and VisTD, the models largely fail to learn any reasonable policy within the allotted time.
It remains an open question whether the complex decision process associated with manual stopping Touchdown navigation is tractable using RL, without any supervised trajectories.

\section{Collection of Human Expert Trajectories}
\label{app:experts}
The collection of human expert trajectories for purposes of establishing a performance upper bound is not very time-intensive.
A player (paid 20\$ per hour) who is familiar with text adventure games played through all five environments to collect the trajectories.
Depending on the environment, the expert spent up to 30 minutes familiarizing themselves with the environment, then played approximately 50 episodes per environment, which are recorded to established human expert performance.
During human playthroughs, the player is subject to the same step count limit as the RL agent.
The maximum step count limit is 64 steps (for Touchdown), hence each episode is relatively quick in terms of play time.

For RTFM, Messenger, and NetHack, the human player observes a symbolic rendering of the grid along with a key that describes which symbol means which entity.
The text is rendered below the grid.
The human player then types in the command they would like to execute.
For ALFWorld, the player observes the text rendering of the scene, as well as a list of text commands to choose from.
The player then types in the index of the command they would like to execute.
For Touchdown, the player observes a colour-coded rendering of the segmentation mask (of the panorama the player is in).
x-coordinates are provided along the bottom of the segmentations, and a list of x-coordinates that the agent may advance towards at the next step is also provided.
The player than chooses the index of the direction they would like to proceed in.

\begin{figure}
  \centering
  \begin{minipage}[b]{\textwidth}
      \includegraphics[width=\linewidth]{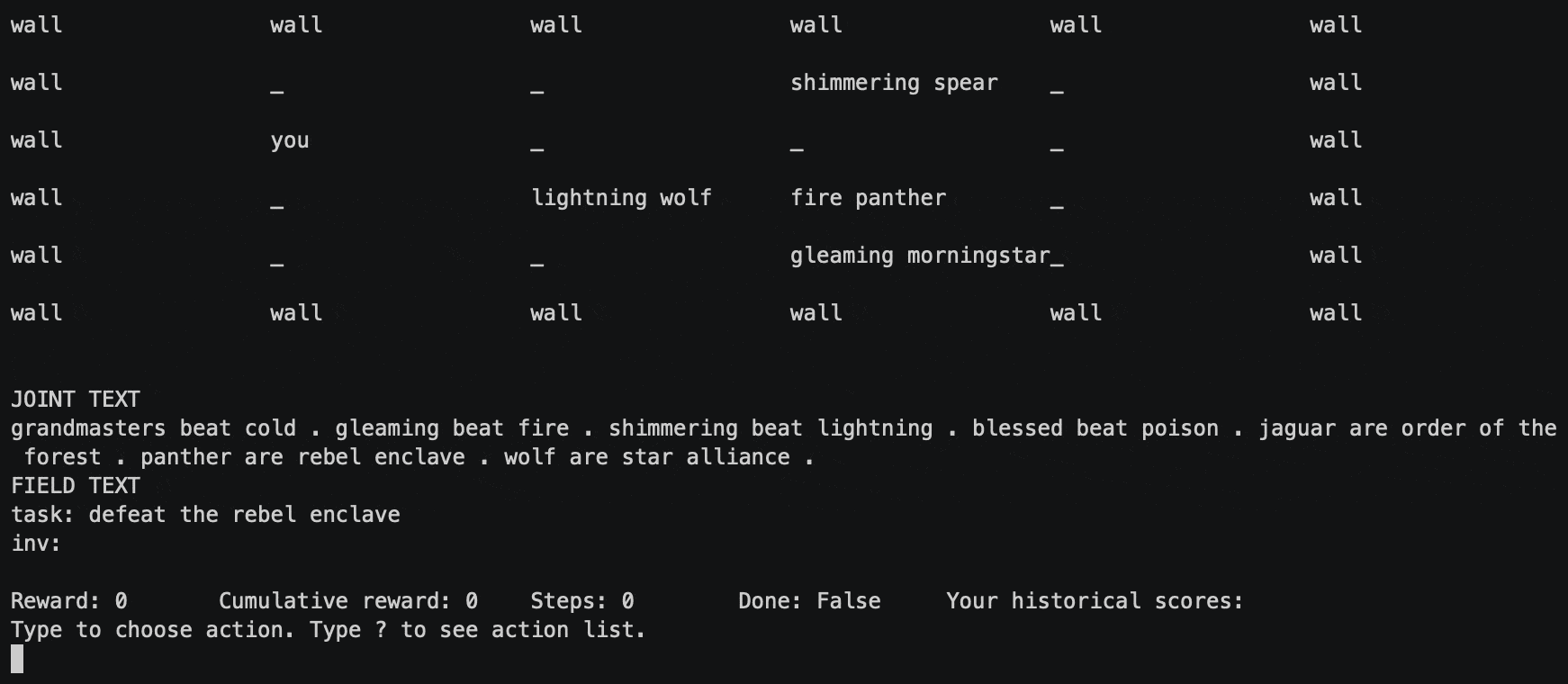}
      \caption{Play interface for RTFM.}
      \label{fig:play_rtfm}
  \end{minipage}
  \begin{minipage}[b]{\textwidth}
    \includegraphics[width=\linewidth]{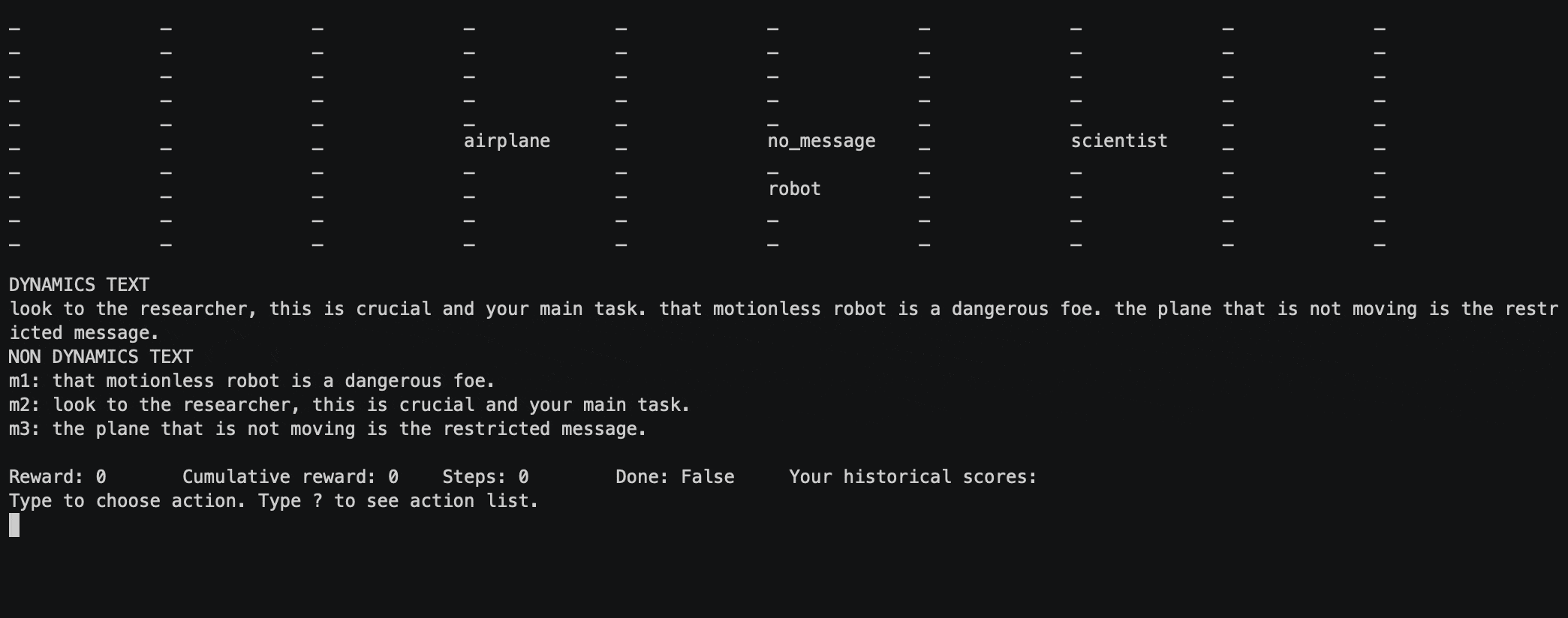}
    \caption{Play interface for Messenger.}
    \label{fig:play_msgr}
  \end{minipage}
\end{figure}

\begin{figure}
  \centering
  \begin{minipage}[b]{0.51\textwidth}
    \includegraphics[width=\linewidth]{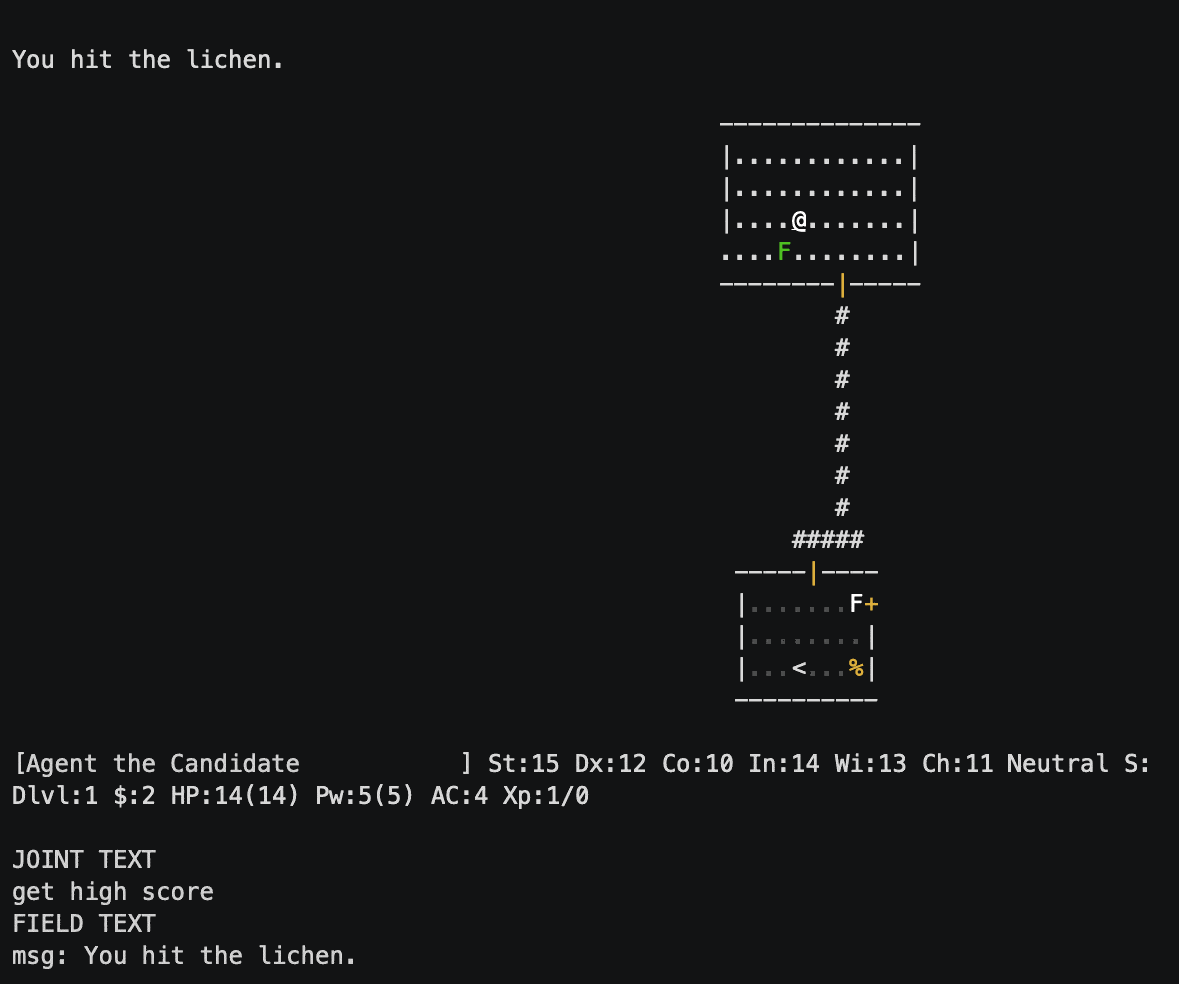}
    \caption{Play interface for NetHack.}
    \label{fig:play_nethack}
  \end{minipage}
  \begin{minipage}[b]{0.47\textwidth}
    \includegraphics[width=\linewidth]{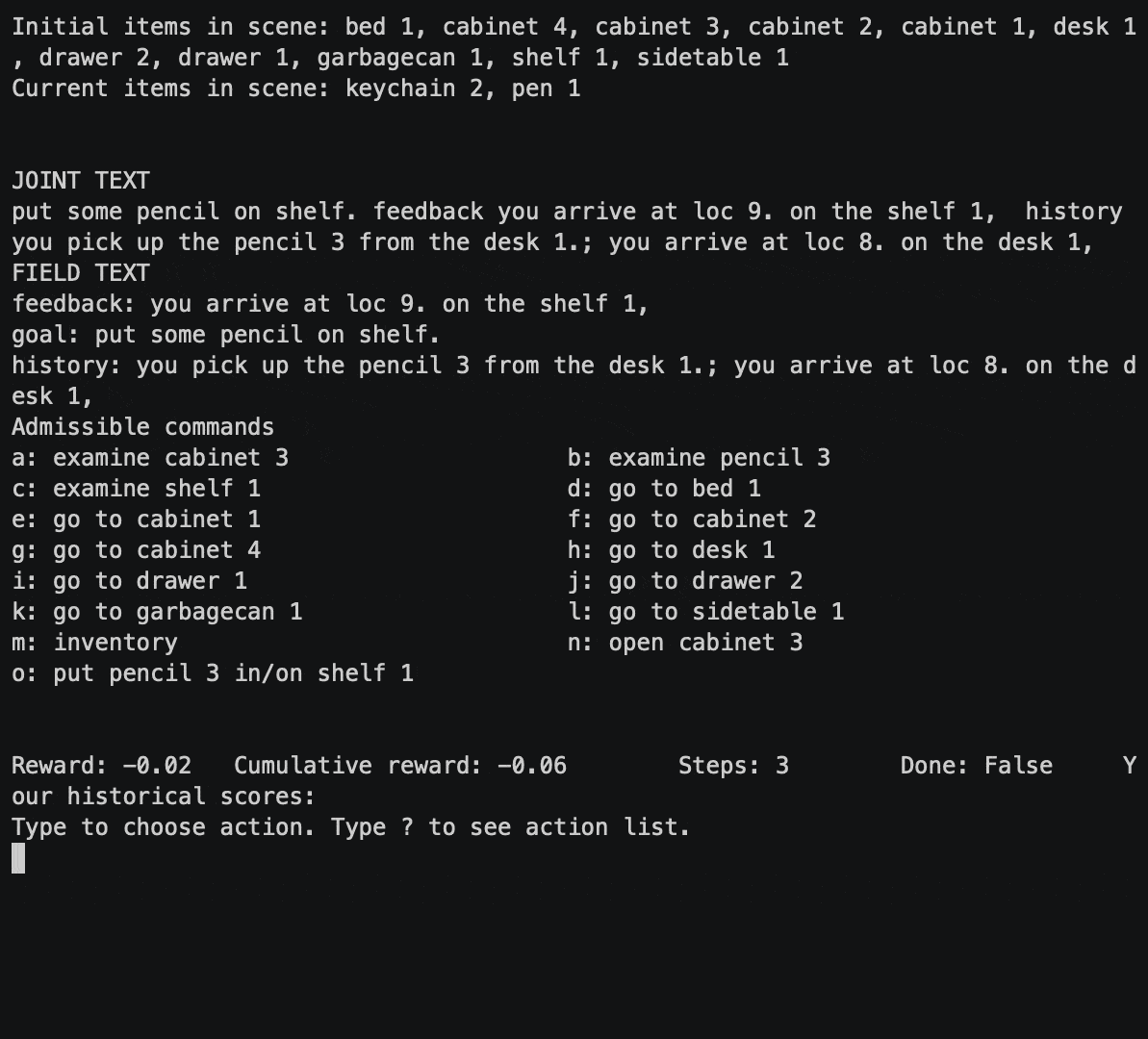}
    \caption{Play interface for ALFWorld.}
    \label{fig:play_alfworld}
  \end{minipage}
\end{figure}

\begin{figure}
    \centering
    \includegraphics[width=\linewidth]{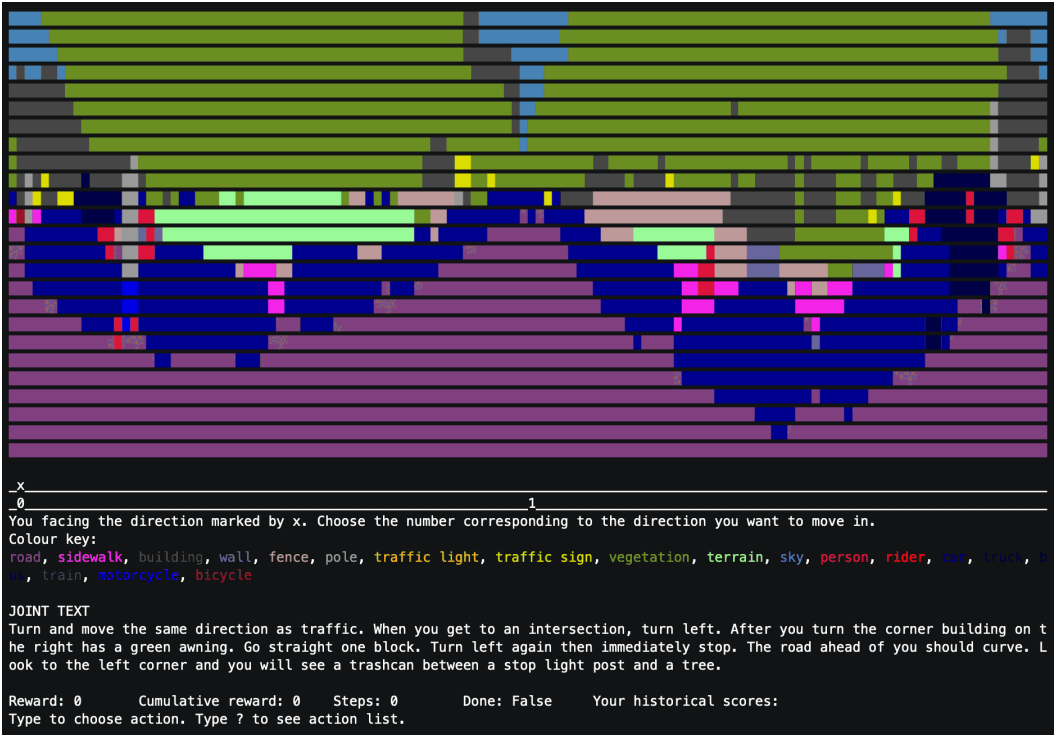}
    \caption{Play interface for VisTD.}
    \label{fig:play_td_vis}
\end{figure}

Playthrough interfaces for RTFM, Messenger, NetHack, ALFWorld, and SymTD are shown in Figure~\ref{fig:play_rtfm} through~\ref{fig:play_td_vis}. 
Figure~\ref{fig:segmentation}~shows examples of segmentation maps that the human player sees playing SymTD.
Unfortunately we cannot include a figure of VisTD due to licensing agreement.


\section{Licenses}
\label{app:license}

We distribute~\benchmarkname~under a MIT LICENSE, which means that researchers are free to modify and distribute our software.
The environments included in~\benchmarkname~use their own corresponding licenses.
These are
\begin{enumerate}
    \item RTFM: \href{https://github.com/facebookresearch/RTFM/blob/master/LICENSE}{Attribution-NonCommercial 4.0 International}
    \item Messenger: \href{https://github.com/ahjwang/messenger-emma/blob/master/LICENSE}{MIT}
    \item NetHack: \href{https://github.com/facebookresearch/nle/blob/master/LICENSE}{NetHack General Public License}
    \item ALFWorld: \href{https://github.com/alfworld/alfworld/blob/master/LICENSE}{MIT}
    \item Touchdown: \href{https://github.com/lil-lab/touchdown/blob/master/LICENSE.txt}{Creative Commons Attribution 4.0 International}
\end{enumerate}

Of particular interest is Touchdown, whose raw panoramas come from Google Streetview.
Neither we nor the creators of Touchdown distribute the panoramas.
Users should follow instructions at~\url{https://sites.google.com/view/streetlearn/dataset}~to obtain the raw panoramas from Google.

\section{Hyperparameters}
\label{app:hyperparam}

By default, we use embedding size $\demb = 100$, and RNN size $\drnn = 200$.
The final representation $\repfinal$ has size 400.
We use 5~\film~layers.
We train using Torchbeast~\citep{torchbeast2019} with an entropy cost of 0.05, baseline cost of 0.5, discount factor of 0.99, step penalty of -0.02, unroll length 80, and learning rate of 0.0005.
We optimize using RMSProp with an epsilon of 0.01 and alpha 0.99.
For Torchbeast parallelization, we use 30 actors, learner batch size of 24, and 4 learner threads.
To account for long text sequences, we use the Huggingface PruneBERT model fine-tuned and distilled on SQuAD~\citep{sanh2020movement}.

Due to GPU memory constraints, we reduce the model size for some environments.
For NetHack, we use 30 embedding size, 100 RNN size, 8 actors, 8 batch size, and 64 unroll length.
For ALFWorld, we use 10 batch size.
For the Touchdown variants, we use 30 embedding size, 100 RNN size, 200 final representation size, 8 actors, 3 batch size, 64 unroll length, and 3~\film~layers.

\section{Compute resources}
\label{app:compute}
To produce our experiments, we ran 7 models each for RTFM, Messenger, and NetHack.
Moreover, we ran 5 models for ALFWorld, SymTD, VisTD, Manual SymTD, and Manual VisTD.
In total, this resulted in $7\times3 + 5\times5 = 46$ models.
We used 5 seeds for each model, resulting in 230 runs.
Each run required up to 20 CPUs (Intel Xeon) and 1 GPU (NVIDIA Quadro Pascal) for up to 2 weeks on an internal cluster.
In total, we used approximately $20 \times 24 \times 2 \times 230 = 220,800$ CPU hours and $24 \times 2 \times 230 = 11040$ GPU hours.

\end{document}